\documentclass{article} 
\usepackage{iclr2026_conference,times}

\usepackage{amsmath,amsfonts,bm}

\def\eqref#1{equation~\ref{#1}}
\def\Eqref#1{Eq.~(\ref{#1})}

\def\1{\bm{1}}

\DeclareMathAlphabet{\mathsfit}{\encodingdefault}{\sfdefault}{m}{sl}
\SetMathAlphabet{\mathsfit}{bold}{\encodingdefault}{\sfdefault}{bx}{n}

\newcommand{\softmax}{\mathrm{softmax}}

\usepackage[colorlinks]{hyperref} 
\hypersetup{citecolor=[HTML]{2980B9}, linkcolor=black}
\newcommand{\pos}[1]{{\color{blue} #1}}
\newcommand{\negc}[1]{{\color{red!70!black} #1}}

\usepackage{url}
\usepackage{enumitem}
\usepackage{amsmath,amsfonts,amssymb,amsthm}
\usepackage{ifthen}
\usepackage{xcolor}

\usepackage{colortbl}

\usepackage{diagbox}
\usepackage{mathtools}
\usepackage{pifont}
\usepackage{multirow}
\usepackage{multicol}
\usepackage{threeparttable}
\usepackage{comment}
\usepackage{subfigure}
\usepackage{subcaption}

\usepackage{enumitem}
\usepackage{tikz}
\usepackage{pgfplots}
\usepackage[multiple]{footmisc}
\usepackage{bm}
\usepackage{booktabs}
\usepackage{array}
\usepackage{siunitx}
\usepackage{tabularx}
\usepackage{geometry}
\usepackage{lipsum}
\usepackage{arydshln}
\usepackage{mdframed}
\usepackage{listings}

\newmdenv[
  backgroundcolor=gray!5,
  linecolor=gray!40,
  roundcorner=3pt,
  skipabove=6pt,
  skipbelow=6pt
]{promptframe}

\usepackage{contour}
\usepackage{dsfont}
\usepackage{makecell}
\usepackage{booktabs,amssymb,siunitx}

\newcolumntype{R}[1]{>{\raggedleft\arraybackslash}p{#1}}

\usepackage{tikz}
\usepackage{tcolorbox}

\usepackage{wrapfig} 
\usepackage{pifont}
\usepackage{xcolor}
\definecolor{LavenderLight}{HTML}{C7C3F5}

\newcommand{\cmark}{\textcolor{green!60!black}{\ding{51}}} 
\newcommand{\xmark}{\textcolor{red!70!black}{\ding{55}}}   

\usepackage{mypreamble}

\newcommand\TODO[1][]{{\color{orange}[TODO\ifthenelse{\equal{#1}{}}{}{: #1}]}}

\newcommand{\ours}{\textsc{{PESO}}}

\newcommand{\pretrain}{\textsc{{Pretrain}}}
\newcommand{\single}{\textsc{{Single evolving LoRA}}}

\newcommand{\cloraall}{\textsc{SumLoRA}$_{\textsc{all}}$}
\newcommand{\cloralatest}{\textsc{SumLoRA}$_{\textsc{latest}}$}
\newcommand{\cloraallinherit}{\textsc{SumLoRA}$_{\textsc{all+inherit}}$}
\newcommand{\cloralatestinherit}{\textsc{SumLoRA}$_{\textsc{latest+inherit}}$}

\newcommand{\sdloraall}{\textsc{SD-LoRA}$_{\textsc{all}}$}
\newcommand{\sdloralatest}{\textsc{SD-LoRA}$_{\textsc{latest}}$}
\newcommand{\sdloraallinherit}{\textsc{SD-LoRA}$_{\textsc{all+inherit}}$}
\newcommand{\sdloralatestinherit}{\textsc{SD-LoRA}$_{\textsc{latest+inherit}}$}

\newcommand{\infall}{\textsc{InfLoRA}$_{\textsc{all}}$}
\newcommand{\inflatest}{\textsc{InfLoRA}$_{\textsc{latest}}$}
\newcommand{\infallinherit}{\textsc{InfLoRA}$_{\textsc{all+inherit}}$}
\newcommand{\inflatestinherit}{\textsc{InfLoRA}$_{\textsc{latest+inherit}}$}

\title{Continual Low-Rank Adapters for LLM-based Generative Recommender Systems}

\author{
Hyunsik Yoo$^{\dagger}$, Ting-Wei Li$^{\dagger}$, SeongKu Kang$^{\ddagger}$, Zhining Liu$^{\dagger}$,
Charlie Xu$^{\S}$, Qilin Qi$^{\S}$, Hanghang Tong$^{\dagger}$ \\
$^{\dagger}$University of Illinois Urbana-Champaign \quad
$^{\ddagger}$Korea University \quad
$^{\S}$Amazon \\
\texttt{\{hy40, twli, liu326, htong\}@illinois.edu} \quad
\texttt{\{seongkukang\}@korea.ac.kr} \\
\texttt{\{caizhx, qilinqi\}@amazon.com}
}

\iclrfinalcopy 
\begin{document}

\maketitle

\begin{abstract}

While large language models (LLMs) achieve strong performance in recommendation, they face challenges in continual learning as users, items, and user preferences evolve over time. Existing LoRA-based continual methods primarily focus on preserving performance on previous tasks, but this overlooks the unique nature of recommendation: the goal is not to predict past preferences, and outdated preferences can even harm performance when current interests shift significantly. To address this, we propose \ours\ (\underline{P}roximally r\underline{E}gularized \underline{S}ingle evolving l\underline{O}ra), a continual adaptation method for LoRA in recommendation. \ours\ introduces a proximal regularizer that anchors the current adapter to its most recent frozen state, enabling the model to flexibly balance adaptation and preservation, and to better capture recent user behaviors. Theoretically, we show that this proximal design provides data-aware, direction-wise guidance in the LoRA subspace. Empirically, \ours\ consistently outperforms existing LoRA-based continual learning methods.
Our code and documentation are available at \href{https://github.com/hsyoo32/peso}{https://github.com/hsyoo32/peso}.

\end{abstract}

\section{Introduction}
Large language models (LLMs) are increasingly used for recommendation by treating the task as sequence generation: given a user’s interaction history, the model autoregressively generates the next item tokens~\citep{bao2025bi, cao2024aligning, tan2024idgenrec, wang2024learnable, bao2023tallrec, kweon2025uncertainty, lin2025order, lin2026mixture,wei2025cofirec}. In practice, LLM is fine-tuned on user histories paired with their next interactions, aligning it with the recommendation objective. However, real-world interaction data are continuously collected and evolve over time: new users and items appear, and user preferences drift~\citep{li2025haystack, qiu2024tucket}. Periodic retraining from scratch on both historical and new data is possible but highly inefficient, making \textit{continual learning} (i.e., updating the model effectively with new data) a natural and appealing solution.

It is well known that a continual model must balance \textit{stability} (retaining past knowledge) and \textit{plasticity} (adapting to new knowledge)~\citep{zhu2021reliable, arani2022learning, ye2022future, zhang2024incmsr, yuan2021one, do2023continual, mi2020ader, yoo2025continual}.
However, continual recommender systems present unique interpretations of these concepts, and bear subtle but critical difference from other domains such as computer vision \citep{bao2025latte,zeng2026subspace}.
In most other domains, continual tasks are typically disjoint and not time-ordered (e.g., cats vs. dogs → trucks vs. sedans), and the primary objective is to preserve performance on previous tasks (stability) while adapting to new ones (plasticity).
In contrast, the ultimate goal of continual recommendation is to accurately capture evolving user preferences in order to predict which items a user {\em will} prefer in the near future. 
That is, recommendation is not concerned with predicting past user preferences; in fact, outdated preferences can even hinder performance if current user interests have shifted significantly (e.g., a user starts preferring romance over action). Thus, stability in recommendation refers to preserving long-term user preferences (e.g., enduring interests in certain genres or brands) that remain predictive, even if they are not strongly reflected in recent data. 
Plasticity, on the other hand, is required to overwrite outdated preferences and to capture emerging trends. This distinct setting in turn requires careful model design.

A common recipe for fine-tuning LLMs in recommendation is Low-Rank Adaptation (LoRA)~\citep{hu2022lora, liu2025cora, zeng2025hierarchical,qiu2026remix}, 
due to simplicity and modularity across components (e.g., attention layers). LoRA freezes pretrained weights and injects lightweight trainable low-rank matrices. Its efficiency makes LoRA a natural candidate for continual learning, motivating our focus on continual LoRA for LLM-based recommenders. 
A simple intuitive approach is to maintain a \textit{single evolving LoRA}: sequentially fine-tuning one adapter, initializing it from the previous stage and optimizing it on new data. This provides strong plasticity while parameter inheritance provides partial preservation of past knowledge. However, it inevitably overwrites useful past knowledge during fine-tuning, leading to forgetting.

To mitigate forgetting, several works in vision have proposed the family of \textit{cumulative LoRA}~\citep{wu2025sd, liang2024inflora, lu2024adaptive, wang2023orthogonal, liu2024learning}, which typically use the sum of the new trainable adapter and all frozen past adapters. 
This design explicitly enhances stability by reusing prior adapters and expanding LoRA’s effective capacity, and it works well when tasks are largely independent (i.e., with minimal interference), allowing each adapter to encode task-specific knowledge. Intuitively, this might seem beneficial for recommendation, where preserving useful past preferences matters. However, our analysis shows that cumulative LoRA often underperforms the simpler single evolving LoRA. Unlike vision tasks, recommendation involves reappearing users with continuously evolving preferences. The model must therefore capture useful interference across stages, but frozen adapters entangle outdated and relevant preferences, making them hard to disentangle. In addition, as adapters accumulate over time, cumulative LoRA incurs growing storage costs and struggles to reflect their relative importance during aggregation.

To address these limitations, we adopt two principles: (1) avoid multiple adapters, which implicitly assume task independence, and (2) preserve past knowledge in a way that supports understanding of current user behavior.
Guided by this, we propose \ours\ (\underline{P}roximally r\underline{E}gularized \underline{S}ingle evolving l\underline{O}ra), which maintains a single evolving LoRA adapter while regularizing it toward its past state with a lightweight proximal term. 
Unlike cumulative LoRA, \ours\ balances stability and plasticity through the natural competition between the data-fitting loss and the proximal term, allowing the model to decide what to adapt or retain.
Theoretically, we show that this design yields data-aware, direction-wise guidance in the LoRA subspace.
We further instantiate it with a per-module softmax–Kullback–Leibler (KL) proximal, which preserves internal module structure rather than treating all parameters equally (i.e., a more nuanced stability mechanism). Empirically, \ours\ consistently outperforms both cumulative LoRA and the single evolving adapter across multiple real-world datasets, achieving a more effective stability–plasticity balance for recommendation.

In summary, our main contributions are threefold.  \textbf{(1) Analysis:} we identify the distinctive stability–plasticity challenge in continual recommendation and show empirically that cumulative LoRA, while effective in simulated user-disjoint settings, underperforms in the natural case where user preferences evolve across time stages; \textbf{(2) Method and Theory:} we propose \ours, a \textit{proximally regularized LoRA} that anchors each update to the previous state, with theory showing direction-wise, data-aware guidance and a per-module softmax–KL instantiation;
\textbf{(3) Experiments:} we demonstrate through extensive experiments on real-world datasets that \ours\ consistently outperforms both single evolving and cumulative LoRA.

\vspace{-1mm}
\section{Preliminary}
\vspace{-2mm}

\textbf{Notations.}
We consider an LLM-based recommender that, given a user’s interaction history, autoregressively predicts the next item token.
At time stage $t\in\{1,\ldots,T\}$, let $\mathcal U_t$ be the set of active users, $\mathcal I_t$ the item set, and let $\mathcal E_t=\{(x_{u,1},\dots,x_{u,N_u})\}_{u\in\mathcal U_t}$ denote the collection of user interaction sequences, where $x_{u,n}\in\mathcal I_t$.
For notational simplicity, we present stage-$t$ data as one prefix-target pair per user:
\AL{
\mathcal D_t=\{(\mathbf{x}_u,y_u):u\in\mathcal U_t,\ N_u\ge 2\},
\quad
\mathbf{x}_u=(x_{u,1},\dots,x_{u,N_u-1}),
\quad
y_u=x_{u,N_u}.
}
In practice, training often uses multiple sliding-window next-item pairs induced from each sequence (see Appendix~\ref{app:experimental_setup}).
Each item 
is represented by \textit{semantic ID} 
 obtained by a codebook-based tokenizer (e.g., RQ-VAE, \citealp{ rajput2023recommender}) trained on item semantic features (e.g., title/description), yielding fixed number of token IDs for each item. Semantic ID captures hierarchical semantics of items and works well in practice.\footnote{Adapting the tokenizer to new items over time is an interesting direction; here we fix the item tokenizer to isolate continual adaptation of the model (LoRA).}

\textbf{Stability and Plasticity in Continual Recommendation.}
We assume an initial model is pretrained offline on base data $\mathcal D_1$, and then fine-tuned sequentially on chronologically arriving blocks $\mathcal D_2,\dots,\mathcal D_T$.
The goal of continual recommendation is to minimize predictive risk on future interactions by balancing \textit{stability} (retaining persistent long-term preferences) and \textit{plasticity} (adapting to new or shifting preferences from recent data), thereby capturing evolving user interests (see Appendix~\ref{app:evolving_pref} for a formal conceptual model).
Concretely, for $\mathcal D_t$, the LLM is fine-tuned with the standard autoregressive cross-entropy for next-item prediction. Let $y=(y_1,\dots,y_M)$ denote the semantic-ID token sequence of the target item. The stage-$t$ training loss is
\AL{\label{eq:cross_entropy}
L_{\mathrm{ce}}^{\mathcal D_t}
\;=\;
\mathbb{E}_{(x,y)\sim \mathcal D_t}\big[-\log p_\theta(y\mid x)\big],
\qquad
p_\theta(y\mid x)=\prod_{m=1}^{M} p_\theta(y_m\mid x,y_{<m}).
}

\textbf{Low-Rank Adaptation (LoRA).}
LoRA freezes the pretrained LLM weight $W_0 \in \mathbb{R}^{d_{\text{out}}\times d_{\text{in}}}$ and adds a trainable low-rank update:
\vspace{-2mm}
\AL{
\Delta W = 
B A,\quad
A\in\mathbb{R}^{r\times d_{\text{in}}},\ B\in\mathbb{R}^{d_{\text{out}}\times r},\ r\ll \min(d_{\text{in}},d_{\text{out}}),
}
so that for an input $x\in\mathbb{R}^{d_{\text{in}}}$ the layer computes $(W_0+\Delta W)\,x$. 
Only $A$ and $B$ are updated during fine-tuning, while $W_0$ remains fixed. 
This yields substantial parameter savings and modular, layer-wise adaptation (e.g., on attention projections). 
In this work, our analysis and method operate entirely within this LoRA subspace and therefore inherit its efficiency. We now formally define our problem.

\begin{PRB}
(Continual adaptation of a generative recommender)
\textbf{Given:}
(1) a pretrained LLM-based recommendation model (fine-tuned with LoRA on $\CAL D_1$), (2) a sequence of chronological data blocks $\mathcal{D}_2, \dots, \mathcal{D}_T$;
\textbf{Goal:}
learn updates that, at each stage $t$, adapt the model to $\CAL D_t$ 
 while retaining useful knowledge from earlier stages, achieving high quality next-item recommendation via a balanced stability–plasticity.
 \end{PRB}

\section{Analysis of LoRA Variants for Continual Recommendation}\label{sec:cumul_analysis}
We introduce two primary baselines for our problem: single evolving LoRA and the cumulative LoRA family. 
Then, we empirically compare them on a natural chronological split and a user-disjoint split.

\textbf{{Single evolving LoRA.}} 
At stage $t$, the LoRA matrices $A_t$ and $B_t$ are initialized (i.e., parameter inheritance) from the previous stage ($A_{t-1}$ and $B_{t-1}$) and fine-tuned on new data $\mathcal{D}_t$:
\AL{
W_t=W_0+B_tA_t, \qquad B_t{\leftarrow}B_{t-1}, A_t{\leftarrow}A_{t-1}, \qquad t \geq 2,
}
where $W_{0}$ is the pretrained LLM weight (i.e., not LoRA updates). 
This baseline is simple and adapts effectively to new data, while parameter inheritance provides partial preservation of past knowledge at initialization. However, it inevitably overwrites useful past knowledge during fine-tuning, leading to forgetting.

\textbf{{Cumulative LoRA Variants}}.
To mitigate forgetting, cumulative LoRA has been widely used in domains such as vision~\citep{wu2025sd, liang2024inflora}.
At stage $t$, it reuses frozen adapters from past stages and adds a new trainable adapter by summing them during both training and inference. The effective update is
\AL{
    W_t = W_0 + \sum_{i=1}^{t-1}\alpha_i \hat{B}_i\hat{A}_i+B_tA_t, \qquad t\geq 2,
}
where $W_{0}$ is the pretrained LLM weight; $\{\hat{B}_i\}_{i=1}^{t-1}$ and $\{\hat{A}_i\}_{i=1}^{t-1}$
are frozen adapters from previous stages; and $B_t,A_t$ are trainable at stage $t$. 
Following prior practice, we use normalized directions $\hat{B}_i=B_i/\|B_i\|_F$ and $\hat{A}_i=A_i/\|A_i\|_F$, which improves stability. 
The scalar $\alpha_i$ are fixed or learned magnitudes.
This design explicitly enhances stability and expands LoRA’s effective capacity, expected too work well when sequential tasks interfere minimally. However, for recommendation where user preferences evolve, this rationale weakens.
To examine this, we study SumLoRA, which uses simple summation, in four variants: (i) \textit{all}, summing all past adapters; (ii) \textit{latest}, summing only the most recent adapter; (iii) \textit{all+inherit}, summing all past adapters with parameter inheritance; and (iv) \textit{latest+inherit}, using only the latest adapter with parameter inheritance. The \textit{all} variant corresponds to the original design of cumulative LoRA family.
We also consider SD-LoRA, which extends summation with learnable magnitudes, with \textit{all} equivalent to \citet{wu2025sd}. For analysis, we focus on the empirically stronger \textit{latest+inherit}. Table~\ref{tab:cumul_analysis} summarizes these design choices.

\begin{table}[t]
\centering
\caption{(Left) Design choices; (Right) performance gain vs. single evolving LoRA (w.r.t. NDCG@5) in different task settings on Instrument dataset.}
\vspace{-1em}
\small
\resizebox{1.00\textwidth}{!}{
\begin{tabular}{l|ccc| r r r}
\toprule
\multicolumn{1}{l|}{} &
\multicolumn{3}{c|}{\textbf{Design choices}} &
\multicolumn{3}{c}{\textbf{Task settings}} \\
\multicolumn{1}{l|}{\textbf{Method}} & \textbf{Learnable mag.} & \textbf{Only latest} & \textbf{Param inherit} & \textbf{(1) User-disjoint} & \textbf{(2) Natural split} & \textbf{Diff. (1)-(2)} \\ \midrule
\cloraall            & \xmark & \xmark & \xmark & $-8.13\%$  & $-26.77\%$ & \pos{$18.64\%$} \\
\cloralatest         & \xmark & \cmark & \xmark & $-12.20\%$ & $-22.05\%$ & \pos{$9.85\%$}  \\
\cloraallinherit     & \xmark & \xmark & \cmark & $-3.25\%$  & $1.57\%$   & \negc{$-4.82\%$} \\
\cloralatestinherit  & \xmark & \cmark & \cmark & $0.00\%$   & $2.36\%$   & \negc{$-2.36\%$} \\
\sdloralatestinherit & \cmark & \cmark & \cmark & $3.25\%$   & $0.79\%$   & \pos{$2.46\%$}  \\
\bottomrule
\end{tabular}}
\label{tab:cumul_analysis}
\vspace{-2em}
\end{table}

\textbf{{Two settings.}}
We evaluate methods in the two settings derived from the same user-item interaction data of Amazon Review (Musical Instruments) dataset: 
\textbf{(1) Natural chronological split:} Interactions are sorted by time; a large portion (e.g., 60\%) is used for pretraining (i.e., $\mathcal{D}_1$), and the remainder is divided into four equal incremental blocks, yielding ${\mathcal{D}_2,\dots,\mathcal{D}_5}$. For each $\mathcal{D}_t$, we apply leave-one-out per user (second-to-last item for validation, last item for test). See Appendix~\ref{app:experimental_setup} for details.
\textbf{(2) Pseudo user-disjoint split:} Users are randomly partitioned into disjoint sets for $\mathcal{D}_t$ ($t=1,\dots,5$), with block sizes matched to the chronological split.
Item order within each user’s sequence is preserved. While similar users may induce some shared preferences across stages, this setting introduces relatively less cross-stage interference than the natural chronological case.

\noindent\textbf{Results.} Table~\ref{tab:cumul_analysis} reports \textbf{(1)} the relative gain vs.\ single evolving LoRA on the user-disjoint split, \textbf{(2)} the relative gain on the chronological split, and \textbf{(3)} their difference (i.e., \textbf{(1)-(2)}). 
We summarize the findings:
First, the \textbf{Diff.}\ column shows that the original cumulative design (i.e., \cloraall) performs much worse in the natural chronological setting than in the user-disjoint setting, confirming that it is better suited for tasks with minimal interference and ill-suited for recommendation.
Second, in the \textbf{Natural split}, \cloraall\ performs worst, followed by \textit{latest}, \textit{all+inherit}, and \textit{latest+inherit}, suggesting that (a) aggregating all past adapters hinders adaptation, and (b) parameter inheritance is essential for gradual, proximal evolution of LoRA with respect to the previous state.
Finally, \sdloralatestinherit\ fails to improve over fixed-magnitude \cloralatestinherit, since useful past components are entangled with stale ones, making weighting ineffective.
Overall, continual recommendation requires evolving adapters with {\em controlled stability}, rather than rigid reuse of past ones, to capture user preference dynamics.

\section{Proposed Framework: \ours}\label{sec:method}
\begin{figure}[t]
\includegraphics[width=1.0\textwidth]{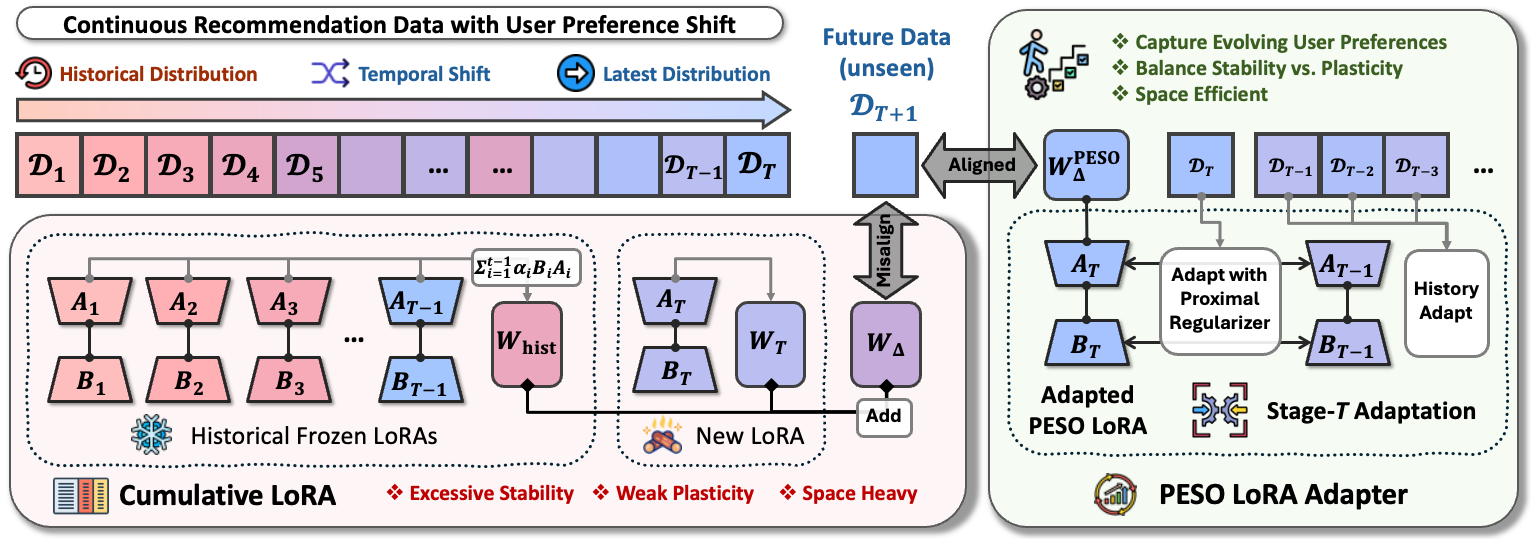}
\vspace{-1.7em}
\caption{Conceptual overview of Cumulative LoRA and our proposed \ours\ with proximal regularizer.}
\label{fig:overview}
\vspace{-1em}
\end{figure}

\vspace{-2mm}
Our design philosophy is to (1) avoid using multiple LoRA adapters, which implicitly assume task independence, and (2) preserve past knowledge in a way that supports understanding of current user behavior. Guided by this, we propose \ours\ (\underline{P}roximally r\underline{E}gularized \underline{S}ingle evolving l\underline{O}ra), which maintains a single evolving LoRA adapter and regularizes each update by keeping the current adapter close to the previous one (shown in Figure~\ref{fig:overview}). 
We begin by presenting the \textit{quadratic proximal framework} and its theoretical implications, and then instantiate \ours\ with a \textit{softmax–KL proximal} to demonstrate its practical effect.

\subsection{Single Evolving LoRA with a Proximal Regularizer}
\label{sec:4.1}
\textbf{General framework.}
We maintain a single evolving LoRA and anchor each update to the previous adapter with a proximal term.
Let $v_t\in\mathbb{R}^m$ denote the concatenation of all flattened LoRA $A/B$ parameters at time stage $t$.
We partition coordinates into groups $g\in\{1,\dots,G\}$ at the level of LoRA factor matrices, so that each group corresponds to the flattened parameters of either $A$ or $B$ for one LoRA-injected module (e.g., attention projections $q/k/v/o$ and MLP projection layers).
We write $v^{(g)}$ for the parameters in group $g$.
The overall loss function for time stage $t$ is
\begin{align}\label{eq:general_framework}
L_t \;=\; L^{\mathcal D_t}_{\mathrm{ce}}
\;+\; \underbrace{\frac{\lambda}{2}\sum_{g=1}^G \|\,v^{(g)}_t - v^{(g)}_{t-1}\,\|_{H^{(g)}_{t-1}}^2}_{\text{proximal term}}, \quad
v_t\leftarrow v_{t-1}\ \text{at init,}
\end{align}
where $L^{\mathcal D_t}_{\mathrm{ce}}$ is the data-fitting term on $\mathcal D_t$ (i.e., cross-entropy, \Eqref{eq:cross_entropy}),
$\|z\|_{H}^2:=z^\top H z$, $\lambda>0$ controls regularization strength, and
each $H^{(g)}_{t-1}\succeq 0$ is a (symmetric) PSD metric that is fixed during stage $t$; 
it can be constant (e.g., $H^{(g)}_{t-1}=I$, corresponding to the L2 case) or precomputed at the previous adapter $v^{(g)}_{t-1}$.
We initialize $v_t\!\leftarrow\! v_{t-1}$ so the proximal penalty starts at zero and grows only as $v_t$ departs from $v_{t-1}$.
This design leverages the natural competition between the data-fitting loss (which pulls toward the optimal state for $\mathcal D_t$) and the proximal term (which pulls toward the previous state).
Next, we theoretically show how this yields data-aware, direction-wise guidance in the LoRA subspace.

\textbf{Theoretical setup.} 
To analyze how the proximal term interacts with the data-fitting loss, we approximate the data-fitting term.
We restrict updates to a fixed $m$-dimensional LoRA subspace.
Let $\theta_0\in\mathbb{R}^d$ be the parameter vector (base LLM and LoRA) after training on the first data block ($t{=}1$). From $t\ge2$, let $\theta(v)=\theta_0+U v$ with $U\in\mathbb{R}^{d\times m}$ and non-LoRA coordinates frozen (i.e., assume $U=[I_m\ 0]$).
For input $x=(\text{prompt},\text{item sequence})$ and next-item token $y$, let $s(\theta,x)$ be the scalar logit of the ground-truth token.
Linearize once at $v=0$:
\begin{align}
s(\theta_0+Uv,x)\;\approx\; s(\theta_0, x)+\Phi(x)^\top v,\qquad
\Phi(x):=U^\top\nabla_\theta s(\theta_0,x)\in\mathbb{R}^m,
\end{align}
where $\Phi(x)$ is a tangent feature of $x$.
For analysis we use a mean-squared-error surrogate for \Eqref{eq:cross_entropy} and define the stage-$t$ optimum
$v_t^*=\arg\min_v L^{\mathcal D_t}(v)$. 
A second-order expansion at $v_t^*$ yields quadratic loss 
\begin{align}\label{eq:quadratic_risk}
L^{\mathcal D_t}(v)\;\approx\;\frac12\,(v-v_t^*)^\top \Sigma_t\,(v-v_t^*),
\qquad
\Sigma_t \;=\; \mathbb{E}_{x\sim \mathcal D_t}\big[\Phi(x)\Phi(x)^\top\big]\succeq 0,
\end{align}
where $\Sigma_t$ is the tangent-feature second-moment matrix for time stage $t$, capturing \textit{how much the stage-$t$ data
supports different directions in the LoRA subspace} (i.e., $u^\top \Sigma_t u=\mathbb{E}_{\mathcal D_t}[(\Phi(x)^\top u)^2]$  $\forall u\in\mathbb{R}^m$).
See Appendix~\ref{sec:appendix_setup} for full setup and assumptions.
In what follows, we present a general proposition showing that our proximal framework yields direction-wise interpolation between the new optimum and the previous adapter, and then derive its L2 corollary to provide intuition into the stability–plasticity balance.

\vspace{1mm}
\begin{PRP}[Generalized–eigen interpolation with a quadratic proximal]
\label{PRP:blockwise-gep}
Let $\Sigma_t=\Sigma_t^\top\succeq 0$.
Define the block-diagonal proximal metric $H_{t-1}:=\operatorname{blkdiag}\!\big(H^{(1)}_{t-1},\dots,H^{(G)}_{t-1}\big)\succeq 0,$ 
with each $H^{(g)}_{t-1}$ symmetric PSD and independent of $v$ during stage $t$.
Under the quadratic approximation in \Eqref{eq:quadratic_risk}, our loss \Eqref{eq:general_framework} is:
\begin{align}]\label{eq:approxloss}
L_t(v)
=\frac12\,(v-v_t^*)^\top \Sigma_t (v-v_t^*)
+\frac{\lambda}{2}\,(v-v_{t-1})^\top H_{t-1} (v-v_{t-1}).
\end{align}
Let $\hat v_t := \arg\min_v L_t(v)$.
Let $\{(q_k,\rho_k)\}_{k=1}^{r}$ be generalized eigenpairs of $(\Sigma_t,H_{t-1})$ on $\mathrm{range}(H_{t-1})$
(i.e., $\Sigma_t q_k=\rho_k H_{t-1} q_k$), normalized by $q_i^\top H_{t-1} q_j=\delta_{ij}$, where $r=\operatorname{rank}(H_{t-1})$.
With $\langle u,w\rangle_{H_{t-1}}:=u^\top H_{t-1} w$,
\begin{align}
\langle \hat v_t ,q_k\rangle_{H_{t-1}}
=\frac{\rho_k}{\rho_k+\lambda}\,\langle v_t^*,q_k\rangle_{H_{t-1}}
+\frac{\lambda}{\rho_k+\lambda}\,\langle v_{t-1},q_k\rangle_{H_{t-1}},
\qquad k=1,\dots,r.
\end{align}
\end{PRP}
The proof of \PRPref{blockwise-gep} is deferred to Appendix~\ref{app:prp1}.
To build intuition, we specialize \PRPref{blockwise-gep} to the \textit{L2 case by taking $H_{t-1}=I$}.
Then the generalized eigenpairs reduce to ordinary eigenpairs of $\Sigma_t$ and
$\langle\cdot,\cdot\rangle_{H_{t-1}}$ becomes the standard inner product, yielding the following corollary.

\vspace{1mm}
\begin{COR}[L2 special case of \PRPref{blockwise-gep}]
\label{COR:l2}
Take $H_{t-1}=I$. If $\Sigma_t q_k=\sigma_k^2 q_k$ with $\{q_k\}$ orthonormal, 
\begin{align}
\langle \hat v_t,q_k\rangle
=\frac{\sigma_k^2}{\sigma_k^2+\lambda}\,\langle v_t^*,q_k\rangle
+\frac{\lambda}{\sigma_k^2+\lambda}\,\langle v_{t-1},q_k\rangle, \qquad k=1,\cdots,m.
\end{align}
\end{COR}
In a nutshell, \CORref{l2} shows a \textbf{data-aware balance between stability and plasticity} in our framework.
Recall that $\Sigma_t=\mathbb{E}_{\mathcal D_t}[\Phi(x)\Phi(x)^\top]$ summarizes how much the stage-$t$ data
\emph{supports} different directions in the LoRA subspace.
Its eigenvectors $q_k$ are principal directions, with eigenvalues $\sigma_k^2$ measuring the strength of support along each direction under $\mathcal D_t$.
By \CORref{l2}, along any $q_k$ the update is a weighted average of $v_t^*$ and $v_{t-1}$,
with weight toward $v_t^*$ equal to $\sigma_k^2/(\sigma_k^2+\lambda)$.
Thus, when $\sigma_k^2$ is large (strong support in $\mathcal D_t$), $\hat v_t$ moves toward $v_t^*$ along $q_k$ (e.g., the user starts engaging more with mystery than sci-fi);
when $\sigma_k^2$ is small (weak support), $\hat v_t$ stays close to $v_{t-1}$ (e.g., a stable brand affinity not observed this week).
If $\sigma_k^2=0$, the component along $q_k$ is kept exactly from the previous stage. 
See Appendix~\ref{app:intuition_on_col2} for a more detailed intuitive explanation.

\subsection{Softmax–KL as a Proximal Regularizer}
\label{sec:4.2}
As shown earlier, the L2 proximal (i.e., $H^{(g)}_{t-1}=I$) is a special case of our general proximal form with $H_{t-1}$. However, it penalizes all coordinate changes equally, treating modules uniformly, ignoring internal structure, and not adapting to the previous state $v_{t-1}$. To address this, we instantiate the proximal term with a \textit{softmax–KL proximal} that preserves per-module structure and leverages the previous state. Formally, the stage-$t$ objective of \ours\ is:
\begin{align}
L_t
\;=\;
L^{\mathcal D_t}_{\mathrm{ce}}
\;+\;\underbrace{
\lambda \sum_{g=1}^G
D_{\mathrm{KL}}\!\big(\softmax(v_t^{(g)})\,\|\,\softmax(v_{t-1}^{(g)})\big)}_{\CAL K_{\textrm{blk}}(v_t,v_{t-1})},
\quad
v_t\leftarrow v_{t-1}\ \text{at init.}
\end{align}
We first show that the softmax–KL proximal locally reduces to a quadratic form, and then give a corollary that interprets it as a $p$-weighted variance, providing an intuitive view of its module-wise stability.

\vspace{1mm}
\begin{PRP}[Per-module softmax–KL is locally quadratic]
\label{PRP:main-kl-local}
Let $v_t^{(g)}$ be the subvector for group $g\in\{1,\dots,G\}$ (e.g., LoRA $A$/$B$ of a module),
$p^{(g)}=\softmax(v_{t-1}^{(g)})$, and $\Delta^{(g)}=v_t^{(g)}-v_{t-1}^{(g)}$. For small $\Delta^{(g)}$,
\begin{align}
&\quad \mathcal{K}_{\mathrm{blk}}(v_t,v_{t-1})
=\frac{\lambda}{2} \sum_{g=1}^G \big(\Delta^{(g)}\big)^\top
\!\Big(\mathrm{diag}\big(p^{(g)}\big)-p^{(g)}{p^{(g)}}^\top\Big)\Delta^{(g)}
+ o(\|\Delta\|^2) \\
&=\frac{\lambda}{2}\,\Delta^\top
\underbrace{\mathrm{blkdiag}\big(H^{(1)}_{t-1},\dots,H^{(G)}_{t-1}\big)}_{=:H_{t-1}}
\,\Delta
+ o(\|\Delta\|^2),
\text{ with } H^{(g)}_{t-1}=\mathrm{diag}(p^{(g)})-p^{(g)}(p^{(g)})^\top\succeq 0.   \notag
\end{align}
\end{PRP}
The proof of \PRPref{main-kl-local} is deferred to Appendix~\ref{app:kl-local}.
\PRPref{main-kl-local} shows the \textbf{softmax–KL proximal is locally equivalent to the quadratic form} $\frac{\lambda}{2}\|v_t-v_{t-1}\|_{H_{t-1}}^2$ with
$H_{t-1}=\mathrm{blkdiag}(H^{(1)}_{t-1},\dots,H^{(G)}_{t-1})$.
Hence, \PRPref{blockwise-gep} applies directly, suggesting a data-aware balance of stability and plasticity.

\vspace{1mm}
\begin{COR}[Softmax–KL equals $p$-weighted variance]
\label{COR:blockwise-kl-variance}
With notation as above, up to an additive constant,
\begin{align}
\mathcal{K}_{\mathrm{blk}}(v_t,v_{t-1})
=
\frac{\lambda}{2}\sum_{g=1}^G \mathrm{Var}_{p^{(g)}}\!\big(\Delta^{(g)}\big),
\end{align}
where $\mathrm{Var}_{p^{(g)}}(\Delta^{(g)})=\sum_{i\in g} p_i^{(g)}(\Delta_i^{(g)}-\mu^{(g)})^2$ and
$\mu^{(g)}=\sum_{i\in g} p_i^{(g)}\Delta_i^{(g)}$.
\end{COR}
\CORref{blockwise-kl-variance} shows that the softmax--KL proximal can be interpreted as a \textit{$p$-weighted variance} of parameter changes.
Consequently, it penalizes (1) \textit{reshuffling}
(i.e., relative (centered) changes) within each module's LoRA factor, and (2) deviations more strongly for coordinates with higher prior mass.
\textbf{This yields module-wise, previous-state–aware stability} without killing plasticity: updates still move toward new optima where data provides strong support (as in \PRPref{blockwise-gep}), while staying close to the previous state otherwise.

\section{Experiments}\label{sec:experiments}
\vspace{-2mm}
We design experiments to answer four key questions: \textbf{RQ1:} To what extent does \ours\ outperform competitors? \textbf{RQ2:} Which proximal regularizer works best in \ours? \textbf{RQ3}: How effectively does \ours\ balance stability and plasticity under different user drift patterns? \textbf{RQ4:} How do hyperparameters affect performance of \ours? \textbf{RQ5}: How does \ours\ compare to traditional continual recommenders?

\textit{Please refer to Appendix~\ref{app:experiments} for additional experimental results and analyses, including distribution-shift quantification, comparisons with additional baselines (full-parameter fine-tuning, O-LoRA, AM-LoRA, LSAT etc.), experiments with other backbones and datasets, and an efficiency analysis.}


\subsection{Experimental Settings}\label{sec:experimental_settings}

\textbf{Datasets.}
We use the real-world Amazon Review dataset, which contains user reviews (treated as implicit interactions) on products over time.
We focus on three categories: Musical Instruments, Movies \& TV, and Books.\footnote{\href{https://amazon-reviews-2023.github.io/}{https://amazon-reviews-2023.github.io/}} 
Detailed preprocessing steps and dataset statistics are provided in Appendix~\ref{app:experimental_setup}. The processed data yield $\{\mathcal{D}_1,\dots,\mathcal{D}_5\}$, where $\mathcal{D}_1$ is a large pretraining set and ${\mathcal{D}_2,\dots,\mathcal{D}_4}$ are smaller incremental sets.

\textbf{Evaluation.}
For each $\mathcal{D}_t$, we apply leave-one-out evaluation per user, reserving the last item for testing. 
Following~\citep{wang2024learnable, bao2025bi}, we construct multiple training pairs $(x_u,y_u)$ per user using a sliding window of size 20. 
Starting from the LLM pretrained on $\mathcal{D}_1$, at each stage $t=2,\dots,5$ the model is fine-tuned and then generates 10 items via constrained beam search restricted to valid item tokens. 
We report Hit@5/10 and NDCG@5/10, averaged over $\mathcal{D}_2,\dots,\mathcal{D}_4$. Full evaluation details are in Appendix~\ref{app:experimental_setup}.

\textbf{Compared methods and implementation details.}
We compare PESO with several LoRA-based baselines for continual learning, all using the same cross-entropy loss and Llama-3.2 1B \citep{grattafiori2024llama} as backbone. The bottom baseline is \pretrain, trained on $\mathcal{D}_1$ and directly evaluated at $t=2,\dots,4$. Among continual methods, we consider: (1) \textit{single evolving LoRA}; and (2) the \textit{cumulative family}, which combines past and current adapters: \textit{SumLoRA}, \textit{SD-LoRA} \citep{wu2025sd}, and \textit{InfLoRA} \citep{liang2024inflora}. SD-LoRA learns magnitudes for normalized past adapters, while InfLoRA precomputes LoRA-$A$ via SVD of the input covariance and trains only $B$, to better align with current data and reduce task-interference. 
As discussed in Section~\ref{sec:cumul_analysis},
original cumulative designs use \textit{all past adapters without inheritance} (\textit{all}). For recommendation, we further test three variants: \textit{latest} (most recent only), \textit{all+inherit} (all with inheritance), and \textit{latest+inherit} (latest with inheritance).
For hyperparameters, $\lambda$ is searched over $[0.5,1.0,2.0,5.0,8.0]$ (set to $2.0$ for Instruments, $5.0$ for Movies\&TV and Books). SD-LoRA magnitudes start at $1.0$.

\begin{table}[t]
  \centering
  \small
  \caption{Recommendation performance averaged across time stages for \ours\ and continual competitors. The best and second-best results are marked in \textbf{bold} and \underline{underline}, respectively. 
  }
\vspace{-1em}
  \resizebox{\linewidth}{!}{
\begin{tabular}{lcccccccccccc}
\toprule
\multicolumn{1}{c}{\textbf{}} &
\multicolumn{4}{c}{\textbf{Instruments}} &
\multicolumn{4}{c}{\textbf{Movies \& TVs}} &
\multicolumn{4}{c}{\textbf{Books}} \\
\multicolumn{1}{l}{\textbf{Methods}}  & 
\textbf{H@5} & \textbf{H@10} & \textbf{N@5} & \textbf{N@10} &
\textbf{H@5} & \textbf{H@10} & \textbf{N@5} & \textbf{N@10} &
\textbf{H@5} & \textbf{H@10} & \textbf{N@5} & \textbf{N@10} \\
\midrule\addlinespace[-0.000ex]
\rowcolor{red!5}
\textsc{Pretrain} & 0.0166 & 0.0216 & 0.0115 & 0.0131 & 0.0166 & 0.0231 & 0.0111 & 0.0132 & 0.0258 & 0.0283 & 0.0196 & 0.0204 \\ 
[-0.4ex]\midrule\addlinespace[-0.000ex]
\rowcolor{orange!7}
\textsc{Single evolving LoRA} & 0.0181 & 0.0253 & 0.0127 & 0.0150 & \underline{0.0175} & \underline{0.0247} & \underline{0.0116} & \underline{0.0138} & \textbf{0.0448} & \underline{0.0557} & \underline{0.0308} & \underline{0.0344} \\ 
[-0.4ex]\midrule\addlinespace[-0.000ex]
\rowcolor{yellow!5}
Cumulative LoRA Family & & & & & & & & & & & & \\
\hdashline
\rowcolor{yellow!10}
\infall            & 0.0156 & 0.0214 & 0.0105 & 0.0124 & 0.0103 & 0.0139 & 0.0067 & 0.0079 & 0.0236 & 0.0332 & 0.0161 & 0.0193 \\
\rowcolor{yellow!10}
\inflatest        & 0.0131 & 0.0167 & 0.0090 & 0.0102 & 0.0073 & 0.0092 & 0.0047 & 0.0054 &   0.0152   &  0.0197      &    0.0108    &   0.0123       \\
\rowcolor{yellow!10}
\infallinherit     & 0.0149 & 0.0219 & 0.0104 & 0.0126 & 0.0109 & 0.0147 & 0.0072 & 0.0085 & 0.0249 & 0.0324 & 0.0171 & 0.0195 \\
\rowcolor{yellow!10}
\inflatestinherit  & 0.0137 & 0.0202 & 0.0095 & 0.0116 & 0.0094 & 0.0132 & 0.0060 & 0.0072 & 0.0225 & 0.0288 & 0.0153 & 0.0174 \\
\hdashline
\rowcolor{green!5}
\cloraall & 0.0134 & 0.0215 & 0.0093 & 0.0119 & 0.0102 & 0.0130 & 0.0067 & 0.0076 & 0.0264 & 0.0402 & 0.0176 & 0.0221 \\
\rowcolor{green!5}
\cloralatest & 0.0143 & 0.0221 & 0.0099 & 0.0124 & 0.0102 & 0.0130 & 0.0067 & 0.0076 & 0.0246 & 0.0354 & 0.0161 & 0.0196 \\
\rowcolor{green!5}
\cloraallinherit & 0.0182 & \underline{0.0260} & 0.0129 & \underline{0.0154} & 0.0160 & 0.0234 & 0.0107 & 0.0131 & 0.0409 & 0.0514 & 0.0287 & 0.0321 \\
\rowcolor{green!5}
\cloralatestinherit & \underline{0.0185} & 0.0255 & \underline{0.0130} & 0.0152 & 0.0172 & 0.0237 & 0.0114 & 0.0135 & 0.0433 & 0.0542 & 0.0306 & 0.0341 \\
\hdashline
\rowcolor{LavenderLight!20}
\sdloraall & 0.0156 & 0.0226 & 0.0107 & 0.0129 & 0.0094 & 0.0133 & 0.0061 & 0.0074 & 0.0238 & 0.0351 & 0.0162 & 0.0198 \\
\rowcolor{LavenderLight!20}
\sdloralatest & 0.0156 & 0.0218 & 0.0102 & 0.0123 & 0.0101 & 0.0142 & 0.0069 & 0.0082 & 0.0241 & 0.0327 & 0.0159 & 0.0186 \\
\rowcolor{LavenderLight!20}
\sdloraallinherit & 0.0176 & 0.0238 & 0.0124 & 0.0144 & 0.0118 & 0.0171 & 0.0077 & 0.0094 & 0.0332 & 0.0412 & 0.0234 & 0.0260 \\
\rowcolor{LavenderLight!20}
\sdloralatestinherit & 0.0184 & 0.0254 & 0.0128 & 0.0150 & 0.0165 & 0.0235 & 0.0109 & 0.0131 & 0.0432 & 0.0530 & \underline{0.0308} & 0.0340 \\
[-0.4ex]\midrule\addlinespace[0.000ex]
\rowcolor{gray!20} \textbf{\ours} & \textbf{0.0193} & \textbf{0.0268} & \textbf{0.0138} & \textbf{0.0162} & \textbf{0.0180} & \textbf{0.0251} & \textbf{0.0118} & \textbf{0.0141} & \textbf{0.0448} & \textbf{0.0569} & \textbf{0.0311} & \textbf{0.0351} \\
[-0.4ex]\midrule\addlinespace[-0.000ex]
\rowcolor{gray!10}\textbf{Performance Gain} (\%) & & & & & & & & & & & &\\
\hdashline
\rowcolor{gray!5}\textsc{vs.} \single & 6.63\% & 5.93\% & 8.66\% & 8.00\% & 2.86\% & 1.62\% & 1.72\% & 2.17\% & 0.00\% & 2.15\% & 0.97\% & 2.03\% \\
\rowcolor{gray!5}
\textsc{vs.} \cloralatestinherit & 4.32\% & 5.10\% & 6.15\% & 6.58\% & 4.65\% & 5.91\% & 3.51\% & 4.44\% & 3.46\% & 4.98\% & 1.63\% & 2.93\% \\
\rowcolor{gray!5}
\textsc{vs.} \sdloralatestinherit & 4.89\% & 5.51\% & 7.81\% & 8.00\% & 9.09\% & 6.81\% & 8.26\% & 7.63\% & 3.70\% & 7.36\% & 0.97\% & 3.24\% \\
\bottomrule
\end{tabular}}
\label{tab:main}
\vspace{-1.5em}
\end{table}



\subsection{Experimental Results and Discussion}\label{sec:exp_results}

\textbf{Main Results (RQ1).}
Table~\ref{tab:main} reports results across four metrics and three datasets in continual settings. 
First, all continual learning methods consistently outperform \pretrain, highlighting the importance of adapting to new data to capture evolving user preferences, even when incremental data is much smaller (e.g., 10\%) than the pretraining data.
Second, neither single evolving LoRA nor the cumulative family dominates, while \ours\ consistently achieves the best results, with average gains of 3.71\%, 4.62\%, and 6.26\% over the best competitors (\single, \cloralatestinherit, and \sdloralatestinherit).
Cumulative LoRA, though more complex and storage-heavy, often underperforms or only matches single evolving LoRA, as rigidly reusing frozen adapters overly constrains adaptation to evolving user preferences. By contrast, \ours\ uses flexible proximal regularization toward the latest state, allowing the data-fitting loss and proximal term to jointly decide what to preserve or update.
Third, as discussed in detail in Section~\ref{sec:cumul_analysis}, regarding SumLoRA and SD-LoRA, original cumulative designs (using all past adapters without inheritance) perform worst, while variants with inheritance or only the latest adapter do better. Notably, some non-inheritance variants even fall below \pretrain, showing that without gradual evolution, continual learning can harm more than help.
InfLoRA yields the weakest results overall, likely because, although it incorporates input data covariance information, freezing $A$ prevents inheritance and gradual adaptation across time, both of which are crucial in continual recommendation.

\begin{figure}[t]
  \centering
  \begin{minipage}{0.48\textwidth}
    \centering
    \includegraphics[width=\linewidth]{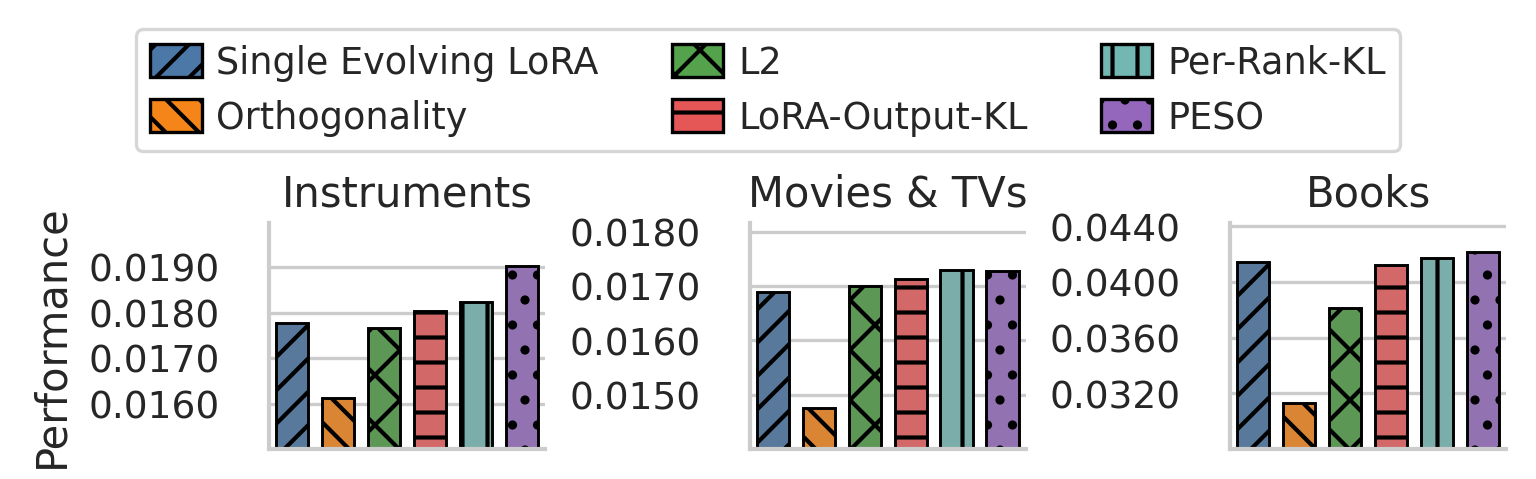}
    \caption{Performance comparison of different regularization methods against the previous LoRA.}
    \label{fig:proximal}
  \end{minipage}\hfill
  \begin{minipage}{0.49\textwidth}
    \centering
    \includegraphics[width=\linewidth]{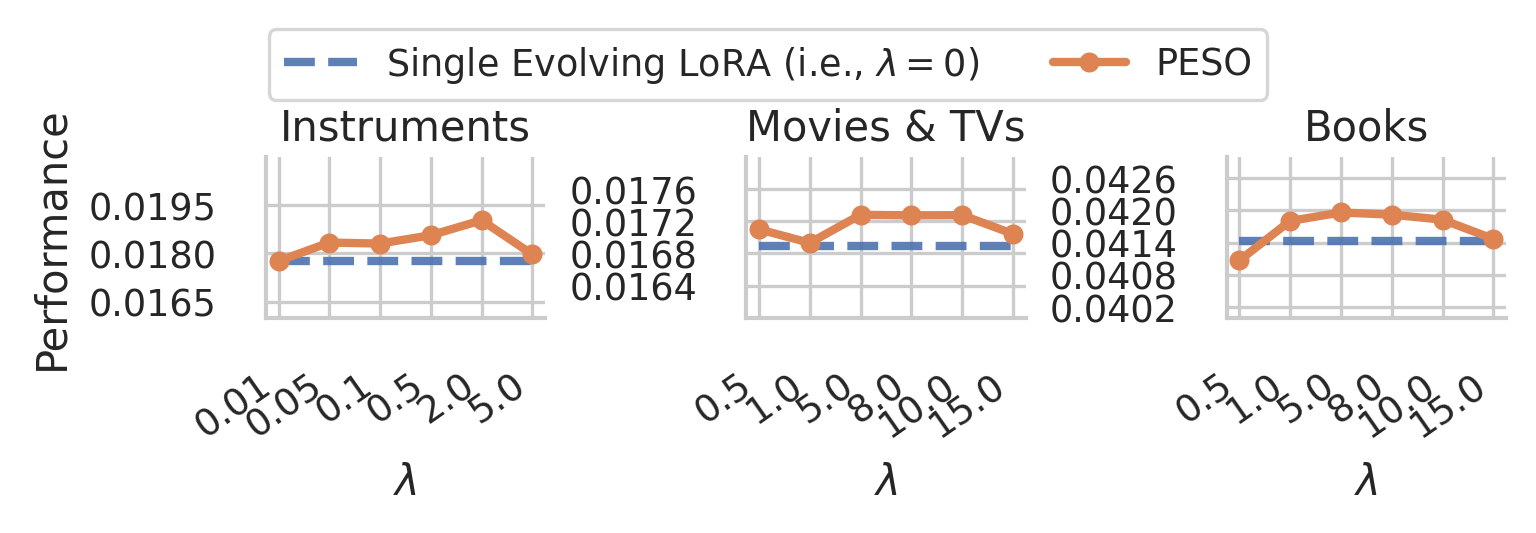}
    \caption{Impact of the scaling weight $\lambda$ for the proximal term on \ours\ performance.}
    \label{fig:lambda}
  \end{minipage}
  \vspace{-2em}
\end{figure}

\textbf{Analysis on Proximal Regularizer (RQ2).}
Unless otherwise noted, all subsequent subsections report average performance across four metrics (Hit@5, Hit@10, NDCG@5, NDCG@10). 
We compare \ours\ with four alternative regularizers on the previous adapter: orthogonality, L2 proximal, LoRA-Output KL, and Per-Rank KL (Figure~\ref{fig:proximal}). Orthogonality, an interference-minimization strategy common in vision, performs far worse than all methods, showing that minimizing interference across stages is harmful in continual recommendation. 
L2 proximal, which penalizes the L2 distance between current and previous parameters, is often comparable to single evolving LoRA but worse than \ours, suggesting that uniform constraints are insufficient. LoRA-Output KL (softmax-KL applied in LoRA output, i.e., function space) and Per-Rank KL (softmax-KL applied on each rank of LoRA matrcies, i.e., finer parameter granularity) are slightly worse or comparable to \ours, suggesting that regularization directly in the parameter space with module-aware structure is more effective, or at least sufficient, compared to output-level or overly fine-grained constraints.

\begin{wraptable}{r}{0.5\columnwidth} 
\vspace{-\intextsep}
\caption{Performance (NDCG@5) across user groups representing stability and plasticity tests.}
\label{tab:user_groups2}
\vspace{-2mm}
\centering
\small
\setlength{\tabcolsep}{3pt}
\renewcommand{\arraystretch}{1.0}
\resizebox{\linewidth}{!}{%
\begin{tabular}{lcc}
\toprule
Method & Dormant Users & New Users \\
\midrule
Single Evolving LoRA (Plasticity) & 0.0154 & 0.0116 \\
Cumulative LoRA (Stability) & 0.0164 & 0.0101 \\
\textbf{\ours\ (Balanced)} & \textbf{0.0170} & \textbf{0.0122} \\
\bottomrule
\end{tabular}%
}
\vspace{-2mm}
\end{wraptable}
\textbf{Stability--Plasticity Analysis via User Groups (RQ3).}
To examine how \ours\ balances long-term interests with newly evolved preferences, we evaluate the final model on two user groups in the Instruments dataset, which serve as proxies for different drift patterns:
(1) \textbf{Dormant Users:} Users who were active in earlier blocks, absent in intermediate blocks, and return in $D_4$. This tests \textbf{stability} (retention of long-term preferences).
(2) \textbf{New Users:} Users who appear only in $D_4$. This tests \textbf{plasticity} (adaptation to new signals).
Table~\ref{tab:user_groups2} illustrates the trade-off: Single Evolving LoRA excels on New Users but performs poorly on Dormant Users due to forgetting, whereas Cumulative LoRA (i.e., \cloralatestinherit) preserves stability but adapts less effectively to New Users. \ours\ achieves the best performance on both groups, indicating a strong stability--plasticity balance.

\textbf{Hyperparameter Analysis (RQ4).}
\textbf{(a) Scaling parameter $\lambda$ for proximal term in \ours.}
Figure~\ref{fig:lambda} shows performance as $\lambda$ varies. Starting from $\lambda=0$ (i.e., single evolving LoRA), performance improves as $\lambda$ increases, then either decreases or plateaus, confirming that $\lambda$ serves as a tunable trade-off between stability and plasticity: too small harms stability; too large harms plasticity. In addition, performance is not highly sensitive to $\lambda$, as results remain stable across a broad range of values.
\textbf{(b) Learning rate for continual stages.}
See Appendix~\ref{app:learning_rate} for full results and discussion.
Since incremental datasets are much smaller than the pretraining set, performance is highly sensitive to learning rate. 
Our results show that using the pretraining rate leads to overfitting, while scaling the rate down ($\approx 0.05–0.1\times $) yields the best performance. 


\begin{wraptable}{r}{0.4\textwidth} 
\vspace{-\intextsep}                  
\caption{Comparison of traditional and LLM-based methods.}
\label{tab:traditional_small}
\vspace{-2mm}
\centering
\small
\resizebox{\linewidth}{!}{%
\begin{tabular}{l|ccc}
\toprule
\textbf{Method} & \textbf{Instruments} & \textbf{Movies \& TVs} & \textbf{Books} \\ \midrule
Pretrain              & 0.0153 & 0.0028 & 0.0041 \\
Fine-tuning           & 0.0180 & 0.0114 & 0.0218 \\
PISA                  & 0.0194 & 0.0106 & 0.0301 \\ \midrule
Pretrain              & 0.0157 & 0.0160 & 0.0235 \\
Fine-tuning   & 0.0178 & 0.0169 & 0.0414 \\
PESO                  & 0.0190 & 0.0173 & 0.0422 \\
\bottomrule
\end{tabular}}
\end{wraptable}

\textbf{Comparison with Traditional Continual Recommenders (RQ5).}
Details are in Appendix~\ref{app:traditional}; Table~\ref{tab:traditional_small} shows a subset (top: traditional, bottom: LLM-based). 
LLM-based methods generally outperform traditional two-tower models, except on Instruments, where explicit dual modeling of users and items helps. 
While \ours\ achieves higher absolute performance, continual methods like PISA~\citep{yoo2025embracing} yield larger relative gains in two-tower models, reflecting the advantage of explicit user embeddings in capturing preference drift and the challenge of doing so with LLMs.

\vspace{-2mm}
\section{Related Works}
\vspace{-2mm}
\textbf{LLM-based Generative Recommender Systems.} 
Recent advances in large language models (LLMs) have inspired generative approaches to recommendation, where the task is framed as sequence generation. Instead of ranking items from a candidate set, the model autoregressively generates the next item token given a user’s interaction history. 
Variants of this paradigm includes zero-shot prompting~\citep{lyu2023llm},
ID-token generation~\citep{tan2024idgenrec, wang2024learnable}, 
data-efficient fine-tuning~\citep{lin2024data}, 
uncertainty-aware decoding~\citep{kweon2025uncertainty}, and alignment techniques for recommendation objectives~\citep{cao2024aligning, bao2025bi, chen2024softmax}. These works demonstrate that LLMs can flexibly leverage textual and structural signals for recommendation, but they typically assume static data. In contrast, real-world interactions arrive continuously, requiring models that can adapt to evolving user preferences without costly retraining. Our work addresses this gap by studying continual adaptation of generative LLM recommenders.

\textbf{Continual Learning for Foundational Models and LoRA.}
Classical continual recommenders use parameter regularization~\citep{ wang2021graph, wang2023structure, yoo2024ensuring}, replay buffers~\citep{ahrabian2021structure, zhang2024influential, zhu2023reloop2, zou2025transformer}, or dynamic architectures~\citep{he2023dynamically, zhang2023continual, lee2025capturing}. With large foundational models, parameter-efficient fine-tuning (PEFT) has become central, with LoRA~\citep{hu2022lora} as a standard choice.
In vision, several continual extensions have been proposed, such as cumulative aggregation of frozen adapters~\citep{liang2024inflora, lu2024adaptive} and learnable magnitude scaling (SD-LoRA)~\citep{wu2025sd}, which are effective when tasks interference is minimal. However, these methods are less suitable for recommendation, where user preferences evolve over time. Our work differs by proposing a proximal single evolving LoRA that avoids the forgetting of single evolving LoRA and the rigidity of cumulative LoRA, better suiting the continual recommendation setting.

\vspace{-1mm}
\section{Conclusion}
\vspace{-2mm}
We have studied the problem of continual adaptation for LLM-based generative recommender systems, where user interactions arrive over time and preferences evolve. Single evolving LoRA offers strong plasticity but suffers from forgetting, while cumulative LoRA improves stability but entangles outdated signals. Our proposed PESO strikes a better balance by maintaining a single adapter and regularizing it toward its prior state, allowing the model to decide what to adapt and what to preserve.
Our theoretical analysis has shown that the proximal design provides data-aware, direction-wise guidance in the LoRA subspace, and our instantiation with per-module softmax–KL further preserves internal parameter structure. Empirical results across multiple real-world datasets confirm that PESO consistently outperforms existing baselines, achieving a superior stability–plasticity balance.
Future directions include drift-aware and more personalized regularization, as well as integrating our approach with other parameter-efficient tuning methods to further improve continual adaptation in large foundation models.


\newpage

\section*{Acknowledgements}
This work is supported by NSF (2324770) and AFOSR (FA9550-24-1-0002).
The content of the information in this document does not necessarily reflect the position or the policy of the Government, and no official endorsement should be inferred. The U.S. Government is authorized to reproduce and distribute reprints for Government purposes notwithstanding any copyright notation here on.

\section*{Ethics Statement.}
This work focuses on continual learning methods for large language model (LLM)-based recommender systems. It does not involve human subjects, sensitive personal data, or private user information. All experiments are conducted on publicly available benchmark datasets (Amazon Reviews). We followed standard preprocessing protocols, and no personally identifiable information was used or released. While recommender systems can influence user exposure to content, this study is purely methodological and does not deploy or interact with real users. We acknowledge the potential societal risks of recommendation technologies, such as reinforcing biases or filter bubbles, and we emphasize that our method (PESO) is designed as a modular continual learning technique, independent of any particular application domain or societal factors. 

\section*{Reproducibility Statement.}
The paper provides: (1) detailed descriptions of datasets, preprocessing steps, and evaluation protocols (Section~\ref{sec:experimental_settings}, Appendix~\ref{app:experimental_setup}); (2) clear definitions of baselines, the proposed method (PESO), and its theoretical analyses (Sections~\ref{sec:cumul_analysis}, \ref{sec:method}, Appendix~\ref{app:theory}); and (3) hyperparameter settings, search ranges, and sensitivity analyses (Section~\ref{sec:experiments}). Results are reported across multiple datasets and metrics for robustness. Full proofs are included in Appendix~\ref{app:theory}. We will release our implementation and data-processing scripts upon publication to ensure reproducibility.


\newpage 

\bibliography{iclr2026_conference}

@inproceedings{liang2024inflora,
  title={Inflora: Interference-free low-rank adaptation for continual learning},
  author={Liang, Yan-Shuo and Li, Wu-Jun},
  booktitle={Proceedings of the IEEE/CVF Conference on Computer Vision and Pattern Recognition},
  pages={23638--23647},
  year={2024}
}

@article{wu2025sd,
  title={Sd-lora: Scalable decoupled low-rank adaptation for class incremental learning},
  author={Wu, Yichen and Piao, Hongming and Huang, Long-Kai and Wang, Renzhen and Li, Wanhua and Pfister, Hanspeter and Meng, Deyu and Ma, Kede and Wei, Ying},
  journal={arXiv preprint arXiv:2501.13198},
  year={2025}
}

@article{hu2022lora,
  title={Lora: Low-rank adaptation of large language models.},
  author={Hu, Edward J and Shen, Yelong and Wallis, Phillip and Allen-Zhu, Zeyuan and Li, Yuanzhi and Wang, Shean and Wang, Lu and Chen, Weizhu and others},
  journal={ICLR},
  volume={1},
  number={2},
  pages={3},
  year={2022}
}

@inproceedings{liu2025cora,
  title={Cora: Collaborative information perception by large language model’s weights for recommendation},
  author={Liu, Yuting and Zhang, Jinghao and Dang, Yizhou and Liang, Yuliang and Liu, Qiang and Guo, Guibing and Zhao, Jianzhe and Wang, Xingwei},
  booktitle={Proceedings of the AAAI Conference on Artificial Intelligence},
  volume={39},
  number={12},
  pages={12246--12254},
  year={2025}
}

@article{lu2024adaptive,
  title={Adaptive rank, reduced forgetting: Knowledge retention in continual learning vision-language models with dynamic rank-selective lora},
  author={Lu, Haodong and Zhao, Chongyang and Xue, Jason and Yao, Lina and Moore, Kristen and Gong, Dong},
  journal={arXiv preprint arXiv:2412.01004},
  year={2024}
}

@article{bao2025bi,
  title={A bi-step grounding paradigm for large language models in recommendation systems},
  author={Bao, Keqin and Zhang, Jizhi and Wang, Wenjie and Zhang, Yang and Yang, Zhengyi and Luo, Yanchen and Chen, Chong and Feng, Fuli and Tian, Qi},
  journal={ACM Transactions on Recommender Systems},
  volume={3},
  number={4},
  pages={1--27},
  year={2025},
  publisher={ACM New York, NY}
}

@article{cao2024aligning,
  title={Aligning large language models with recommendation knowledge},
  author={Cao, Yuwei and Mehta, Nikhil and Yi, Xinyang and Keshavan, Raghunandan and Heldt, Lukasz and Hong, Lichan and Chi, Ed H and Sathiamoorthy, Maheswaran},
  journal={arXiv preprint arXiv:2404.00245},
  year={2024}
}

@inproceedings{tan2024idgenrec,
  title={Idgenrec: Llm-recsys alignment with textual id learning},
  author={Tan, Juntao and Xu, Shuyuan and Hua, Wenyue and Ge, Yingqiang and Li, Zelong and Zhang, Yongfeng},
  booktitle={Proceedings of the 47th international ACM SIGIR conference on research and development in information retrieval},
  pages={355--364},
  year={2024}
}

@inproceedings{wang2024learnable,
  title={Learnable item tokenization for generative recommendation},
  author={Wang, Wenjie and Bao, Honghui and Lin, Xinyu and Zhang, Jizhi and Li, Yongqi and Feng, Fuli and Ng, See-Kiong and Chua, Tat-Seng},
  booktitle={Proceedings of the 33rd ACM International Conference on Information and Knowledge Management},
  pages={2400--2409},
  year={2024}
}

@article{rajput2023recommender,
  title={Recommender systems with generative retrieval},
  author={Rajput, Shashank and Mehta, Nikhil and Singh, Anima and Hulikal Keshavan, Raghunandan and Vu, Trung and Heldt, Lukasz and Hong, Lichan and Tay, Yi and Tran, Vinh and Samost, Jonah and others},
  journal={Advances in Neural Information Processing Systems},
  volume={36},
  pages={10299--10315},
  year={2023}
}

@article{chen2024softmax,
  title={On softmax direct preference optimization for recommendation},
  author={Chen, Yuxin and Tan, Junfei and Zhang, An and Yang, Zhengyi and Sheng, Leheng and Zhang, Enzhi and Wang, Xiang and Chua, Tat-Seng},
  journal={Advances in Neural Information Processing Systems},
  volume={37},
  pages={27463--27489},
  year={2024}
}

@inproceedings{bao2023tallrec,
  title={Tallrec: An effective and efficient tuning framework to align large language model with recommendation},
  author={Bao, Keqin and Zhang, Jizhi and Zhang, Yang and Wang, Wenjie and Feng, Fuli and He, Xiangnan},
  booktitle={Proceedings of the 17th ACM conference on recommender systems},
  pages={1007--1014},
  year={2023}
}

@inproceedings{kweon2025uncertainty,
  title={Uncertainty Quantification and Decomposition for LLM-based Recommendation},
  author={Kweon, Wonbin and Jang, Sanghwan and Kang, SeongKu and Yu, Hwanjo},
  booktitle={Proceedings of the ACM on Web Conference 2025},
  pages={4889--4901},
  year={2025}
}

@inproceedings{lin2025order,
  title={Order-agnostic Identifier for Large Language Model-based Generative Recommendation},
  author={Lin, Xinyu and Shi, Haihan and Wang, Wenjie and Feng, Fuli and Wang, Qifan and Ng, See-Kiong and Chua, Tat-Seng},
  booktitle={Proceedings of the 48th international ACM SIGIR conference on research and development in information retrieval},
  pages={1923--1933},
  year={2025}
}

@inproceedings{wang2021graph,
  title={Graph structure aware contrastive knowledge distillation for incremental learning in recommender systems},
  author={Wang, Yuening and Zhang, Yingxue and Coates, Mark},
  booktitle={Proceedings of the 30th ACM International Conference on Information \& Knowledge Management},
  pages={3518--3522},
  year={2021}
}

@inproceedings{wang2023structure,
  title={Structure aware incremental learning with personalized imitation weights for recommender systems},
  author={Wang, Yuening and Zhang, Yingxue and Valkanas, Antonios and Tang, Ruiming and Ma, Chen and Hao, Jianye and Coates, Mark},
  booktitle={Proceedings of the AAAI Conference on Artificial Intelligence},
  volume={37},
  number={4},
  pages={4711--4719},
  year={2023}
}

@inproceedings{ahrabian2021structure,
  title={Structure aware experience replay for incremental learning in graph-based recommender systems},
  author={Ahrabian, Kian and Xu, Yishi and Zhang, Yingxue and Wu, Jiapeng and Wang, Yuening and Coates, Mark},
  booktitle={Proceedings of the 30th ACM International Conference on Information \& Knowledge Management},
  pages={2832--2836},
  year={2021}
}

@inproceedings{he2023dynamically,
  title={Dynamically expandable graph convolution for streaming recommendation},
  author={He, Bowei and He, Xu and Zhang, Yingxue and Tang, Ruiming and Ma, Chen},
  booktitle={Proceedings of the ACM Web Conference 2023},
  pages={1457--1467},
  year={2023}
}

@inproceedings{zhang2024influential,
  title={Influential Exemplar Replay for Incremental Learning in Recommender Systems},
  author={Zhang, Xinni and Chen, Yankai and Ma, Chenhao and Fang, Yixiang and King, Irwin},
  booktitle={Proceedings of the AAAI Conference on Artificial Intelligence},
  volume={38},
  number={8},
  pages={9368--9376},
  year={2024}
}

@inproceedings{do2023continual,
  title={Continual Collaborative Filtering Through Gradient Alignment},
  author={Do, Jaime Hieu and Lauw, Hady W},
  booktitle={Proceedings of the 17th ACM Conference on Recommender Systems},
  pages={1133--1138},
  year={2023}
}

@article{arani2022learning,
  title={Learning fast, learning slow: A general continual learning method based on complementary learning system},
  author={Arani, Elahe and Sarfraz, Fahad and Zonooz, Bahram},
  journal={arXiv preprint arXiv:2201.12604},
  year={2022}
}

@article{zhu2021reliable,
  title={Reliable adversarial distillation with unreliable teachers},
  author={Zhu, Jianing and Yao, Jiangchao and Han, Bo and Zhang, Jingfeng and Liu, Tongliang and Niu, Gang and Zhou, Jingren and Xu, Jianliang and Yang, Hongxia},
  journal={arXiv preprint arXiv:2106.04928},
  year={2021}
}

@inproceedings{yoo2025embracing,
  title={Embracing plasticity: Balancing stability and plasticity in continual recommender systems},
  author={Yoo, Hyunsik and Kang, SeongKu and Qiu, Ruizhong and Xu, Charlie and Wang, Fei and Tong, Hanghang},
  booktitle={Proceedings of the 48th International ACM SIGIR conference on research and development in Information Retrieval},
  pages={2092--2101},
  year={2025}
}

@inproceedings{lin2024data,
  title={Data-efficient Fine-tuning for LLM-based Recommendation},
  author={Lin, Xinyu and Wang, Wenjie and Li, Yongqi and Yang, Shuo and Feng, Fuli and Wei, Yinwei and Chua, Tat-Seng},
  booktitle={Proceedings of the 47th International ACM SIGIR Conference on Research and Development in Information Retrieval},
  pages={365--374},
  year={2024}
}

@inproceedings{zhu2023reloop2,
  title={ReLoop2: Building Self-Adaptive Recommendation Models via Responsive Error Compensation Loop},
  author={Zhu, Jieming and Cai, Guohao and Huang, Junjie and Dong, Zhenhua and Tang, Ruiming and Zhang, Weinan},
  booktitle={Proceedings of the 29th ACM SIGKDD Conference on Knowledge Discovery and Data Mining},
  pages={5728--5738},
  year={2023}
}

@inproceedings{mi2020ader,
  title={Ader: Adaptively distilled exemplar replay towards continual learning for session-based recommendation},
  author={Mi, Fei and Lin, Xiaoyu and Faltings, Boi},
  booktitle={Proceedings of the 14th ACM Conference on Recommender Systems},
  pages={408--413},
  year={2020}
}

@inproceedings{qiu2025ask,
title={Ask, and it shall be given: {On} the {Turing} completeness of prompting},
author={Qiu, Ruizhong and Xu, Zhe and Bao, Wenxuan and Tong, Hanghang},
booktitle={The Thirteenth International Conference on Learning Representations},
year={2025},
}

@inproceedings{qiu2025efficient2,
title={How efficient is {LLM-generated} code? {A} rigorous \& high-standard benchmark},
author={Qiu, Ruizhong and Zeng, Weiliang Will and Ezick, James and Lott, Christopher and Tong, Hanghang},
booktitle={The Thirteenth International Conference on Learning Representations},
year={2025},
}

@inproceedings{li2025graph,
title={Graph data selection for domain adaptation: {A} model-free approach},
author={Li, Ting-Wei and Qiu, Ruizhong and Tong, Hanghang},
booktitle={Advances in Neural Information Processing Systems 38},
year={2025},
}

@inproceedings{zou2025transformer,
title={Transformer copilot: {Learning} from the mistake log in {LLM} fine-tuning},
author={Zou, Jiaru and Ban, Yikun and Li, Zihao and Qi, Yunzhe and Qiu, Ruizhong and Yang, Ling and He, Jingrui},
booktitle={Advances in Neural Information Processing Systems 38},
year={2025},
}

@inproceedings{xu2024discrete,
title={Discrete-state continuous-time diffusion for graph generation},
author={Xu, Zhe and Qiu, Ruizhong and Chen, Yuzhong and Chen, Huiyuan and Fan, Xiran and Pan, Menghai and Zeng, Zhichen and Das, Mahashweta and Tong, Hanghang},
booktitle={Advances in Neural Information Processing Systems 37},
year={2024},
}

@inproceedings{liu2025breaking,
title={Breaking silos: {Adaptive} model fusion unlocks better time series forecasting},
author={Liu, Zhining and Yang, Ze and Lin, Xiao and Qiu, Ruizhong and Wei, Tianxin and Zhu, Yada and Hamann, Hendrik and He, Jingrui and Tong, Hanghang},
booktitle={Proceedings of the 42nd International Conference on Machine Learning},
year={2025},
}

@inproceedings{liu2024class,
title={Class-imbalanced graph learning without class rebalancing},
author={Liu, Zhining and Qiu, Ruizhong and Zeng, Zhichen and Yoo, Hyunsik and Zhou, David and Xu, Zhe and Zhu, Yada and Weldemariam, Kommy and He, Jingrui and Tong, Hanghang},
booktitle={Proceedings of the 41st International Conference on Machine Learning},
year={2024},
}

@inproceedings{qiu2024tucket,
title={{TUCKET:} A tensor time series data structure for efficient and accurate factor analysis over time ranges},
author={Qiu, Ruizhong and Jang, Jun-Gi and Lin, Xiao and Liu, Lihui and Tong, Hanghang},
booktitle={Proceedings of the VLDB Endowment 17}, number={13},
year={2024},
}

@inproceedings{bao2025latte,
title={Latte: Collaborative test-time adaptation of vision-language models in federated learning},
author={Bao, Wenxuan and Deng, Ruxi and Qiu, Ruizhong and Wei, Tianxin and Tong, Hanghang and He, Jingrui},
booktitle={Proceedings of the IEEE/CVF International Conference on Computer Vision},
year={2025},
}

@inproceedings{lin2026mixture,
title={Mixture of sequence: {Theme-aware} mixture-of-experts for long-sequence recommendation},
author={Lin, Xiao and Tang, Zhicheng and Cong, Weilin and Hang, Mengyue and Wang, Kai and Wang, Yajuan and Zeng, Zhichen and Li, Ting-Wei and Yoo, Hyunsik and Liu, Zhining and Ning, Xuying and Qiu, Ruizhong and Chen, Wen-Yen and Chang, Shuo and Jin, Rong and Li, Huayu and Tong, Hanghang},
booktitle={Proceedings of the ACM Web Conference 2026},
year={2026},
}

@article{zeng2026pave,
title={Pave your own path: {Graph} gradual domain adaptation on fused {Gromov–Wasserstein} geodesics},
author={Zeng, Zhichen and Qiu, Ruizhong and Bao, Wenxuan and Wei, Tianxin and Lin, Xiao and Yan, Yuchen and Abdelzaher, Tarek F. and Han, Jiawei and Tong, Hanghang},
journal={Transactions on Machine Learning Research},
year={2026},
}

@article{qiu2025efficient,
title={Efficient inference scaling for safety assurance},
author={Qiu, Ruizhong and Li, Gaotang and Li, Ting-Wei and Wei, Tianxin and He, Jingrui and Tong, Hanghang},
journal={NeurIPS 2025 Workshop on Vision--Language Model Real-World Deployment},
year={2025},
}

@article{li2025haystack,
title={Haystack engineering: {Context} engineering meets the long-context challenge in large language models},
author={Li, Mufei and Fu, Dongqi and Wang, Limei and Zhang, Si and Zeng, Hanqing and Sancak, Kaan and Qiu, Ruizhong and Wang, Haoyu Peter and He, Xiaoxin and Bresson, Xavier and Xia, Yinglong and Sun, Chonglin and Li, Pan},
journal={NeurIPS 2025 Workshop on Evaluating the Evolving LLM Lifecycle: Benchmarks, Emergent Abilities, and Scaling},
year={2025},
}

@article{qiu2026remix,
title={{ReMix:} Reinforcement routing for mixtures of {LoRAs} in {LLM} finetuning},
author={Qiu, Ruizhong and Zeng, Hanqing and Xia, Yinglong and Meng, Yiwen and Chen, Ren and Feng, Jiarui and Fu, Dongqi and Wang, Qifan and Liu, Jiayi and Xiao, Jun and Fan, Xiangjun and Zhang, Benyu and Li, Hong and Liu, Zhining and Yoo, Hyunsik and Zeng, Zhichen and Wei, Tianxin and Tong, Hanghang},
journal={Under review},
year={2026},
}

@article{wei2026agentic,
title={Agentic reasoning for large language models: {A} survey},
author={Wei, Tianxin and Li, Ting-Wei and Liu, Zhining and Ning, Xuying and Yang, Ze and Zou, Jiaru and Zeng, Zhichen and Qiu, Ruizhong and Lin, Xiao and Fu, Dongqi and Li, Zihao and Ai, Mengting and Zhou, Duo and Bao, Wenxuan and Li, Yunzhe and Li, Gaotang and Qian, Cheng and Wang, Yu and Tang, Xiangru and Xiao, Yin and Fang, Liri and Liu, Hui and Tang, Xianfeng and Zhang, Yuji and Wang, Chi and You, Jiaxuan and Ji, Heng and Tong, Hanghang and He, Jingrui},
journal={arXiv preprint},
year={2026},
}

@article{zeng2026subspace,
title={Subspace alignment for vision-language model test-time adaptation},
author={Zeng, Zhichen and Bao, Wenxuan and Lin, Xiao and Qiu, Ruizhong and Wei, Tianxin and Ning, Xuying and Yan, Yuchen and Luo, Chen and Cheng, Monica Xiao and He, Jingrui and Tong, Hanghang},
journal={arXiv preprint},
year={2026},
}

@article{wei2025cofirec,
title={{CoFiRec:} Coarse-to-fine tokenization for generative recommendation},
author={Wei, Tianxin and Ning, Xuying and Chen, Xuxing and Qiu, Ruizhong and Hou, Yupeng and Xie, Yan and Yang, Shuang and Hua, Zhigang and He, Jingrui},
journal={arXiv preprint},
year={2025},
}

@article{zeng2025hierarchical,
title={Hierarchical {LoRA} {MoE} for efficient {CTR} model scaling},
author={Zeng, Zhichen and Hang, Mengyue and Liu, Xiaolong and Liu, Xiaoyi and Lin, Xiao and Qiu, Ruizhong and Wei, Tianxin and Liu, Zhining and Yuan, Siyang and Yang, Chaofei and Liu, Yiqun and Yin, Hang and Yang, Jiyan and Tong, Hanghang},
journal={arXiv preprint},
year={2025},
}

@inproceedings{zhang2023continual,
  title={Continual learning on dynamic graphs via parameter isolation},
  author={Zhang, Peiyan and Yan, Yuchen and Li, Chaozhuo and Wang, Senzhang and Xie, Xing and Song, Guojie and Kim, Sunghun},
  booktitle={Proceedings of the 46th International ACM SIGIR Conference on Research and Development in Information Retrieval},
  pages={601--611},
  year={2023}
}

@inproceedings{yuan2021one,
  title={One person, one model, one world: Learning continual user representation without forgetting},
  author={Yuan, Fajie and Zhang, Guoxiao and Karatzoglou, Alexandros and Jose, Joemon and Kong, Beibei and Li, Yudong},
  booktitle={Proceedings of the 44th International ACM SIGIR Conference on Research and Development in Information Retrieval},
  pages={696--705},
  year={2021}
}

@inproceedings{zhang2024incmsr,
  title={IncMSR: An Incremental Learning Approach for Multi-Scenario Recommendation},
  author={Zhang, Kexin and Wang, Yichao and Li, Xiu and Tang, Ruiming and Zhang, Rui},
  booktitle={Proceedings of the 17th ACM International Conference on Web Search and Data Mining},
  pages={939--948},
  year={2024}
}

@inproceedings{ye2022future,
  title={Future gradient descent for adapting the temporal shifting data distribution in online recommendation systems},
  author={Ye, Mao and Jiang, Ruichen and Wang, Haoxiang and Choudhary, Dhruv and Du, Xiaocong and Bhushanam, Bhargav and Mokhtari, Aryan and Kejariwal, Arun and Liu, Qiang},
  booktitle={Uncertainty in Artificial Intelligence},
  pages={2256--2266},
  year={2022},
  organization={PMLR}
}

@article{grattafiori2024llama,
  title={The llama 3 herd of models},
  author={Grattafiori, Aaron and Dubey, Abhimanyu and Jauhri, Abhinav and Pandey, Abhinav and Kadian, Abhishek and Al-Dahle, Ahmad and Letman, Aiesha and Mathur, Akhil and Schelten, Alan and Vaughan, Alex and others},
  journal={arXiv preprint arXiv:2407.21783},
  year={2024}
}

@article{lyu2023llm,
  title={Llm-rec: Personalized recommendation via prompting large language models},
  author={Lyu, Hanjia and Jiang, Song and Zeng, Hanqing and Xia, Yinglong and Wang, Qifan and Zhang, Si and Chen, Ren and Leung, Christopher and Tang, Jiajie and Luo, Jiebo},
  journal={arXiv preprint arXiv:2307.15780},
  year={2023}
}

@inproceedings{he2020lightgcn,
  title={Lightgcn: Simplifying and powering graph convolution network for recommendation},
  author={He, Xiangnan and Deng, Kuan and Wang, Xiang and Li, Yan and Zhang, Yongdong and Wang, Meng},
  booktitle={Proceedings of the 43rd International ACM SIGIR conference on research and development in Information Retrieval},
  pages={639--648},
  year={2020}
}

@inproceedings{wang2023orthogonal,
  title={Orthogonal subspace learning for language model continual learning},
  author={Wang, Xiao and Chen, Tianze and Ge, Qiming and Xia, Han and Bao, Rong and Zheng, Rui and Zhang, Qi and Gui, Tao and Huang, Xuan-Jing},
  booktitle={Findings of the Association for Computational Linguistics: EMNLP 2023},
  pages={10658--10671},
  year={2023}
}

@article{liu2024learning,
  title={Learning attentional mixture of loras for language model continual learning},
  author={Liu, Jialin and Wu, Jianhua and Liu, Jie and Duan, Yutai},
  journal={arXiv preprint arXiv:2409.19611},
  year={2024}
}

@inproceedings{shi2024preliminary,
  title={Preliminary study on incremental learning for large language model-based recommender systems},
  author={Shi, Tianhao and Zhang, Yang and Xu, Zhijian and Chen, Chong and Feng, Fuli and He, Xiangnan and Tian, Qi},
  booktitle={Proceedings of the 33rd ACM International Conference on Information and Knowledge Management},
  pages={4051--4055},
  year={2024}
}

@inproceedings{wang2022learning,
  title={Learning to prompt for continual learning},
  author={Wang, Zifeng and Zhang, Zizhao and Lee, Chen-Yu and Zhang, Han and Sun, Ruoxi and Ren, Xiaoqi and Su, Guolong and Perot, Vincent and Dy, Jennifer and Pfister, Tomas},
  booktitle={Proceedings of the IEEE/CVF conference on computer vision and pattern recognition},
  pages={139--149},
  year={2022}
}

@inproceedings{zheng2024adapting,
  title={Adapting large language models by integrating collaborative semantics for recommendation},
  author={Zheng, Bowen and Hou, Yupeng and Lu, Hongyu and Chen, Yu and Zhao, Wayne Xin and Chen, Ming and Wen, Ji-Rong},
  booktitle={2024 IEEE 40th International Conference on Data Engineering (ICDE)},
  pages={1435--1448},
  year={2024},
  organization={IEEE}
}

@inproceedings{yoo2025continual,
  title={Continual Recommender Systems},
  author={Yoo, Hyunsik and Kang, SeongKu and Tong, Hanghang},
  booktitle={Proceedings of the 34th ACM International Conference on Information and Knowledge Management},
  pages={6857--6860},
  year={2025}
}

@inproceedings{yoo2024ensuring,
  title={Ensuring user-side fairness in dynamic recommender systems},
  author={Yoo, Hyunsik and Zeng, Zhichen and Kang, Jian and Qiu, Ruizhong and Zhou, David and Liu, Zhining and Wang, Fei and Xu, Charlie and Chan, Eunice and Tong, Hanghang},
  booktitle={Proceedings of the ACM web conference 2024},
  pages={3667--3678},
  year={2024}
}

@article{lee2025capturing,
  title={Capturing User Interests from Data Streams for Continual Sequential Recommendation},
  author={Lee, Gyuseok and Yoo, Hyunsik and Hwang, Junyoung and Kang, SeongKu and Yu, Hwanjo},
  journal={arXiv preprint arXiv:2506.07466},
  year={2025}
}

@article{bei2026mem,
  title={Mem-gallery: Benchmarking multimodal long-term conversational memory for mllm agents},
  author={Bei, Yuanchen and Wei, Tianxin and Ning, Xuying and Zhao, Yanjun and Liu, Zhining and Lin, Xiao and Zhu, Yada and Hamann, Hendrik and He, Jingrui and Tong, Hanghang},
  journal={arXiv preprint arXiv:2601.03515},
  year={2026}
}

@article{wei2025evo,
  title={Evo-memory: Benchmarking llm agent test-time learning with self-evolving memory},
  author={Wei, Tianxin and Sachdeva, Noveen and Coleman, Benjamin and He, Zhankui and Bei, Yuanchen and Ning, Xuying and Ai, Mengting and Li, Yunzhe and He, Jingrui and Chi, Ed H and others},
  journal={arXiv preprint arXiv:2511.20857},
  year={2025}
}

@inproceedings{ningmc,
  title={MC-Search: Evaluating and Enhancing Multimodal Agentic Search with Structured Long Reasoning Chains},
  author={Ning, Xuying and Fu, Dongqi and Wei, Tianxin and Ai, Mengting and Zou, Jiaru and Li, Ting-Wei and Tong, Hanghang and Zhu, Yada and Hamann, Hendrik and He, Jingrui},
  booktitle={The Fourteenth International Conference on Learning Representations}
}

@article{liu2025seeing,
  title={Seeing but not believing: Probing the disconnect between visual attention and answer correctness in vlms},
  author={Liu, Zhining and Chen, Ziyi and Liu, Hui and Luo, Chen and Tang, Xianfeng and Wang, Suhang and Zeng, Joy and Dai, Zhenwei and Shi, Zhan and Wei, Tianxin and others},
  journal={arXiv preprint arXiv:2510.17771},
  year={2025}
}

@inproceedings{liu2025selfelicit,
  title={Selfelicit: Your language model secretly knows where is the relevant evidence},
  author={Liu, Zhining and Amjad, Rana Ali and Adkathimar, Ravinarayana and Wei, Tianxin and Tong, Hanghang},
  booktitle={Proceedings of the 63rd Annual Meeting of the Association for Computational Linguistics (Volume 1: Long Papers)},
  pages={9153--9173},
  year={2025}
}

@article{li2025language,
  title={Language in the flow of time: Time-series-paired texts weaved into a unified temporal narrative},
  author={Li, Zihao and Lin, Xiao and Liu, Zhining and Zou, Jiaru and Wu, Ziwei and Zheng, Lecheng and Fu, Dongqi and Zhu, Yada and Hamann, Hendrik and Tong, Hanghang and others},
  journal={arXiv preprint arXiv:2502.08942},
  year={2025}
}

@article{lin2024backtime,
  title={Backtime: Backdoor attacks on multivariate time series forecasting},
  author={Lin, Xiao and Liu, Zhining and Fu, Dongqi and Qiu, Ruizhong and Tong, Hanghang},
  journal={Advances in Neural Information Processing Systems},
  volume={37},
  pages={131344--131368},
  year={2024}
}

@article{he2026fedecider,
  title={FeDecider: An LLM-Based Framework for Federated Cross-Domain Recommendation},
  author={He, Xinrui and Li, Ting-Wei and Wei, Tianxin and Ning, Xuying and He, Xinyu and Bao, Wenxuan and Tong, Hanghang and He, Jingrui},
  journal={arXiv preprint arXiv:2602.16034},
  year={2026}
}

@article{lee2025sprint,
  title={SPRINT: Scalable and Predictive Intent Refinement for LLM-Enhanced Session-based Recommendation},
  author={Lee, Gyuseok and Kweon, Wonbin and Yue, Zhenrui and Liu, Yaokun and Liu, Yifan and Yoon, Susik and Wang, Dong and Kang, SeongKu},
  journal={arXiv preprint arXiv:2508.00570},
  year={2025}
}
\bibliographystyle{iclr2026_conference}

\newpage
\appendix

\section{Conceptual Modeling of Evolving User Preferences}\label{app:evolving_pref}
We assume an initial model is pretrained offline on base data $\CAL D_1$, and then fine-tuned sequentially on chronologically arriving blocks $\mathcal{D}_2, \dots, \mathcal{D}_T$.
Let $x_u^{t-1}$ denote $u$'s interaction history available before stage $t$, and let $P_t(y \mid x_u^{t-1})$ be the conditional distribution of the next item $y$ during stage $t$, representing user preferences.
In continual recommendation, these distributions evolve over time, which can be conceptually modeled as
\AL{
P_{t}(y\mid x_u^{t-1}) \;\approx\; \alpha_t \, P_{t-1}(y \mid x_u^{t-1}) \;+\; (1-\alpha_t)\,Q_t(y \mid x_u^{t-1}),
}
where $P_{t-1}$ captures stability (persistent long-term preferences), $Q_t$ captures plasticity (new or shifting preferences estimated from new data), and $\alpha_t\in[0,1]$ controls the balance. 
The goal is to minimize expected risk on upcoming interactions by balancing stability and plasticity.

\section{Detailed Theoretical Analysis}
\label{app:theory}

\subsection{Setup and Assumptions}
\label{sec:appendix_setup}

\begin{ASS}[Parameters and LoRA subspace]
Let $\theta\in\mathbb{R}^d$ denote the vectorized concatenation of all model parameters (base LLM and LoRA).
Let $\theta_0$ be the parameter vector after training on the first data block ($t{=}1$). From $t\ge2$, restrict updates to a fixed $m$-dimensional LoRA subspace spanned by columns of $U\in \BB R^{d\times m}$ and write
\AL{
\theta=\theta_0+ Uv,\qquad v\in\mathbb{R}^m,
}
with all non-LoRA coordinates frozen. Without loss of generality, assume $U=[I_m \,\,0]$, i.e., the LoRA subspace is the first $m$ coordinates.
\end{ASS}

\begin{ASS}[Linearization and tangent features.]
Let $s(\theta,x)\in\mathbb{R}$ be the scalar logit of the ground-truth next item.
We linearize $s$ at $v=0$ (i.e., at $\theta=\theta_0$): 
\AL{
s(\theta_0+Uv,x)\;\approx\; s(\theta_0,x)+v^\top U^\top \nabla_{\theta} s(\theta_0,x)\;=\; s_0(x)+v^\top\Phi(x),
}
with tangent features of $x$
\AL{
\Phi(x):= U^\top\nabla_{\theta_{}} s(\theta_0,x)\in\mathbb{R}^m.
}
\end{ASS}

\begin{ASS}[Data and loss.]
Let $(x,y)\sim\mathcal{D}_t$ be examples in block $t$.
In recommendation, $x=(\text{prompt}
, \text{item sequence})$ and $y\in \CAL V$ is the next-item token. Training typically uses cross-entropy on logits; for analysis, we use a mean-squared-error (MSE) surrogate.
Define the block-$t$ risk
\AL{
L^{\CAL D_t}(v)\;=\;\mathbb{E}_{(x,y)\sim\mathcal{D}_t}\!\left[\tfrac12\big(s(\theta_0+Uv,x)-r_t(x,y)\big)^2\right].
}
where $r_t(x,y)\in\mathbb{R}$ is a calibrated target score for the ground-truth next item.
\end{ASS}
Note that under the linearization, this yields a quadratic risk with positive-semidefinite curvature. All later proofs use only this PSD curvature, not the exact form of $r_t$.

\begin{ASS}[Quadratic form under the linearization.]
Substituting $s(\theta_0+v,x)\;\approx\; s_0(x)\;+\;\Phi(x)^\top v$ gives, up to an additive constant,
\AL{
L^{\CAL D_t}(v)\;=\;b_t^\top v\;+\;\tfrac12\,v^\top \Sigma_t\, v,\qquad
b_t:=\mathbb{E}_{\CAL D_t}\!\big[(s_0(x)-r_t(x,y))\,\Phi(x)\big],\quad
\Sigma_t:=\mathbb{E}_{\CAL D_t}\!\big[\Phi(x)\Phi(x)^\top\big]\succeq 0.
}
Define stage-$t$ optimum $v_t^*$ as any minimizer of $L^{\mathcal D_t}(v)$
\AL{
v_t^*\;=\;\arg\min_{v} L^{\CAL D_t}(v).
}
A second-order Taylor expansion of $L_t^{\CAL D_t}$ at $v_t^*$ gives
\AL{
L^{\CAL D_t}(v)\;=\;L^{\CAL D_t}(v_t^*)\;+\;\underbrace{(\nabla L^{\CAL D_t})^\top(v_t^*)}_{=\,0} (v-v_t^*)\;+\;\tfrac12\,(v-v_t^*)^\top \underbrace{\nabla^2 L^{\CAL D_t}(v_t^*)}_{=\,\Sigma_t}\,(v-v_t^*) .
}
Dropping the constant term, the centered quadratic risk used throughout is
\AL{
 L^{\CAL D_t}(v)\;=\;\tfrac12\,(v-v_t^*)^\top \Sigma_t\,(v-v_t^*),
}
\end{ASS}
where $\Sigma_t$ is the tangent-feature second-moment matrix for time stage $t$, capturing \textit{how much the stage-$t$ data
supports different directions in the LoRA subspace} (i.e., $u^\top \Sigma_t u=\mathbb{E}_{\mathcal D_t}[(\Phi(x)^\top u)^2]$  $\forall u\in\mathbb{R}^m$).
Also note that we fix the linearization at \(\theta_0\): \(s(\theta_0+Uv,x)\approx s_0(x)+\Phi(x)^\top v\) with \(\Phi(x)=U^\top\nabla_\theta s(\theta_0,x)\). Although $\Phi(x)$ is fixed across $t$, the \(\Sigma_t=\mathbb{E}_{\mathcal{D}_t}[\Phi(x)\Phi(x)^\top]\) varies with the data block distribution, so drift is captured via \(\Sigma_t\) and the shifting optimum \(v_t^*\).

\paragraph{Remark: relinearization per block.}
If desired, one may instead relinearize at \(\theta_0+Uv_{t-1}\), replacing \(\Phi(x)\) by
\(\Phi_{t-1}(x)=U^\top\nabla_\theta s(\theta_0+Uv_{t-1},x)\) and
\(\Sigma_t\) by \(\mathbb{E}[\Phi_{t-1}\Phi_{t-1}^\top]\).
All propositions and closed forms carry over with these substitutions; the only change is
that the curvature reflects the anchor \(v_{t-1}\) of the current block.
We found fixed linearization sufficient and notationally lighter.

\subsection{Proof of \PRPref{blockwise-gep}.}\label{app:prp1}
\begin{ASS}[Complementarity (no doubly--flat directions)]\label{ASS:complementarity}
On the LoRA subspace $\mathbb{R}^m$, let $\Sigma_t\succeq 0$ and $H_{t-1}\succeq 0$ be symmetric and fixed with respect to $v$.
Assume
\AL{
\ker(\Sigma_t)\cap \ker(H_{t-1})=\{0\}.
}
Equivalently, for any $x\in\mathbb{R}^m\setminus\{0\}$,
$
x^\top\Sigma_t x>0  \text{ or }  x^\top H_{t-1} x>0.
$
\end{ASS}

\begin{proof}
Recall that $\hat v_t$ denotes the minimizer of $L_t(v)$ in \Eqref{eq:approxloss}.

\textbf{(i) Closed form of the minimizer.}
Differentiating yields
\AL{
\nabla_v L_t(v)=\Sigma_t(v-v_t^*)+\lambda H_{t-1}(v-v_{t-1}).
}
Setting $\nabla_v L_t(\hat v_t)=0$ gives the normal equation
\begin{equation}\label{eq:normal_equation_app}
(\Sigma_t+\lambda H_{t-1})\,\hat v_t=\Sigma_t v_t^*+\lambda H_{t-1}v_{t-1}.
\end{equation}
Under Assumption~\ref{ASS:complementarity}, for any $x\neq 0$,
\AL{
x^\top(\Sigma_t+\lambda H_{t-1})x=x^\top\Sigma_t x+\lambda x^\top H_{t-1}x>0,
}
so $\Sigma_t+\lambda H_{t-1}\succ 0$ and \Eqref{eq:normal_equation_app} has the unique solution
\AL{
\hat v_t=(\Sigma_t+\lambda H_{t-1})^{-1}\big(\Sigma_t v_t^*+\lambda H_{t-1}v_{t-1}\big).
}

\textbf{(ii) Direction-wise interpolation in a generalized eigenbasis.}
Let $\{(q_k,\rho_k)\}_{k=1}^r$ be generalized eigenpairs of $(\Sigma_t,H_{t-1})$ on $\mathrm{range}(H_{t-1})$,
i.e., $\Sigma_t q_k=\rho_k H_{t-1}q_k$, normalized by $q_i^\top H_{t-1}q_j=\delta_{ij}$.
Left-multiply \Eqref{eq:normal_equation_app} by $q_k^\top$ to obtain
\AL{
q_k^\top\Sigma_t \hat v_t+\lambda q_k^\top H_{t-1}\hat v_t
= q_k^\top\Sigma_t v_t^*+\lambda q_k^\top H_{t-1}v_{t-1}.
}
Using the symmetry of $\Sigma_t$ and $H_{t-1}$, together with $\Sigma_t q_k=\rho_k H_{t-1}q_k$ and
$\langle u,w\rangle_{H_{t-1}}:=u^\top H_{t-1}w$, we obtain
\AL{
(\rho_k+\lambda)\langle \hat v_t,q_k\rangle_{H_{t-1}}
=\rho_k\langle v_t^*,q_k\rangle_{H_{t-1}}
+\lambda\langle v_{t-1},q_k\rangle_{H_{t-1}}.
}
Dividing by $\rho_k+\lambda$ yields the claimed interpolation for $k=1,\dots,r$.

\textbf{(iii) Note on directions in $\ker(H_{t-1})$.}
If $H_{t-1}\succ 0$, then $r=m$ and the above holds for all directions.
If $H_{t-1}$ is singular, the statement is posed on $\mathrm{range}(H_{t-1})$; uniqueness of $\hat v_t$ is still guaranteed by Assumption~\ref{ASS:complementarity}, which rules out doubly-flat directions in $\ker(\Sigma_t)\cap\ker(H_{t-1})$.
\end{proof}

\subsection{Intuitive Explanation of \CORref{l2}}
\label{app:intuition_on_col2}
Here, we provide an intuitive explanation of how \ours\ offers data-aware, direction-wise guidance.

\begin{itemize}
    \item \textbf{Semantics of directions ($q_k$): decoupled preference axes.}
    The eigenvectors $q_k$ of the tangent-feature second-moment matrix $\Sigma_t$ represent principal directions in the LoRA subspace. Intuitively, these directions can be viewed as approximately independent latent axes of user behavior (e.g., one axis may correspond to ``sci-fi affinity'' while another reflects ``price sensitivity''). Since the $q_k$'s form an orthogonal basis (in the L2 case), \ours\ can adjust the model along one axis with minimal interference to others, helping preserve long-term knowledge while adapting to new signals.

    \item \textbf{Mechanism ($\sigma_k^2$): signal strength as a gate.}
    The eigenvalue $\sigma_k^2$ measures how strongly the current block $\mathcal D_t$ supports updates along direction $q_k$. \CORref{l2} shows that \ours\ interpolates between the new-data optimum $v_t^*$ and the previous adapter $v_{t-1}$ with weight $\sigma_k^2/(\sigma_k^2+\lambda)$:
    \begin{itemize}
        \item \textbf{Large $\sigma_k^2$ (strong support $\Rightarrow$ plasticity).}
        When $\mathcal D_t$ provides strong evidence along a direction (e.g., the user recently engages heavily with mystery content), the model is encouraged to move toward $v_t^*$ along $q_k$, enabling rapid adaptation to evolving interests.
        \item \textbf{Small $\sigma_k^2$ (weak support $\Rightarrow$ stability).}
        When $\mathcal D_t$ provides little evidence along a direction (e.g., a stable preference that does not appear in recent interactions), updates along $q_k$ are down-weighted and the solution stays close to $v_{t-1}$, preventing long-term interests from being overwritten by weak or noisy signals.
    \end{itemize}
\end{itemize}

\subsection{Proof of \PRPref{main-kl-local}}\label{app:kl-local}

To prove \PRPref{main-kl-local}, we first establish the following proposition for arbitrary $v_t$ and $v_{t-1}$, and then extend it to the blockwise case.

\begin{PRP}[Local quadratic form of softmax-KL proximal]
\label{PRP:kl-specific}
Let $p:=\mathrm{softmax}(v_{t-1})\in\mathbb{R}^n$ and $\Delta:=v_t-v_{t-1}$. 
Define
\begin{align}
\mathcal{K}(\Delta)\ :=\ D_{\mathrm{KL}}\!\big(\mathrm{softmax}(v_{t-1}+\Delta)\ \|\ \mathrm{softmax}(v_{t-1})\big).
\end{align}
Then $\mathcal{K}(0)=0$, $\nabla \mathcal{K}(0)=0$, and the second-order Taylor expansion at $\Delta=0$ is
\begin{align}
\mathcal{K}(\Delta)\ =\ \tfrac{1}{2}\,\Delta^\top\!\big(\mathrm{diag}(p)-pp^\top\big)\,\Delta\ +\ o(\|\Delta\|^2).
\end{align}
Equivalently,
\begin{align}
\mathcal{K}(\Delta)\ =\ \tfrac{1}{2}\,\underbrace{\Big(\sum_{i=1}^n p_i\,(\Delta_i-\mu)^2\Big)}_{\displaystyle \mathrm{Var}_p(\Delta)}\ +\ o(\|\Delta\|^2),
\qquad
\mu:=\sum_{i=1}^n p_i\,\Delta_i.
\end{align}
\end{PRP}

\begin{proof}
Write $r(\Delta):=\mathrm{softmax}(v_{t-1}+\Delta)\in\mathbb{R}^n$ and $p:=r(0)=\mathrm{softmax}(v_{t-1})$.  
By definition,
\begin{align}
\mathcal{K}(\Delta)\ =\ \sum_{i=1}^n r_i(\Delta)\,\log\!\frac{r_i(\Delta)}{p_i}.
\end{align}

(i) At $\Delta=0$ we have $r(0)=p$, so
\begin{align}
\mathcal{K}(0)=\sum_i p_i\log(p_i/p_i)=0.
\end{align}
For the gradient, differentiate using the scalar identity $\frac{d}{dx}[x\log(x/c)] = \log(x/c)+1$:
\begin{align}
\frac{\partial \mathcal{K}}{\partial \Delta_a}
= \sum_{i=1}^n \frac{\partial r_i}{\partial \Delta_a}\,\Big(\log\!\frac{r_i}{p_i}+1\Big).
\end{align}
Evaluating at $\Delta=0$ gives $\log(r_i/p_i)=0$ and hence
\begin{align}
\Big[\nabla \mathcal{K}(0)\Big]_a
= \sum_{i=1}^n \Big[\frac{\partial r_i}{\partial \Delta_a}\Big]_{\Delta=0}
= \frac{\partial}{\partial \Delta_a}\Big(\sum_{i=1}^n r_i(\Delta)\Big)\Big|_{\Delta=0}
= \frac{\partial}{\partial \Delta_a}(1)\ =\ 0,
\end{align}
since softmax outputs sum to one for all $\Delta$.

(ii) Differentiate the gradient once more:
\begin{align}
\frac{\partial^2 \mathcal{K}}{\partial \Delta_a\,\partial \Delta_b}
= \sum_{i=1}^n \frac{\partial^2 r_i}{\partial \Delta_a\,\partial \Delta_b}\,\Big(\log\!\frac{r_i}{p_i}+1\Big)
\ +\ \sum_{i=1}^n \frac{\partial r_i}{\partial \Delta_a}\,\frac{1}{r_i}\,\frac{\partial r_i}{\partial \Delta_b}.
\end{align}
At $\Delta=0$, the first sum becomes $\sum_i \partial^2 r_i/\partial \Delta_a\partial \Delta_b$ (since $\log(r_i/p_i)=0$), which is zero because $\sum_i r_i(\Delta)\equiv 1$ for all $\Delta$. Thus,
\begin{align}
\Big[\nabla^2 \mathcal{K}(0)\Big]_{ab}
= \sum_{i=1}^n \frac{1}{p_i}\,
\Big[\frac{\partial r_i}{\partial \Delta_a}\Big]_{\Delta=0}\,
\Big[\frac{\partial r_i}{\partial \Delta_b}\Big]_{\Delta=0}.
\end{align}
It remains to compute the Jacobian of softmax at $v_{t-1}$:
\begin{align}
J_{ia}\ :=\ \Big[\frac{\partial r_i}{\partial \Delta_a}\Big]_{\Delta=0}
= \frac{\partial}{\partial v_a}\Big(\frac{e^{v_i}}{\sum_{j} e^{v_j}}\Big)\Big|_{v=v_{t-1}}
= p_i\,(\mathbf{1}\{i=a\}-p_a).
\end{align}
Therefore,
\begin{align}
\Big[\nabla^2 \mathcal{K}(0)\Big]_{ab}
= \sum_{i=1}^n \frac{1}{p_i}\,J_{ia}\,J_{ib}
= \sum_{i=1}^n p_i\,(\mathbf{1}\{i=a\}-p_a)(\mathbf{1}\{i=b\}-p_b).
\end{align}
Expanding the sum gives
\begin{align}
\sum_{i} p_i\,\mathbf{1}\{i=a\}\,\mathbf{1}\{i=b\}
- p_b\sum_{i} p_i\,\mathbf{1}\{i=a\}
- p_a\sum_{i} p_i\,\mathbf{1}\{i=b\}
+ p_ap_b\sum_{i} p_i.
\end{align}
Since $\sum_i p_i=1$ and $\sum_i p_i\,\mathbf{1}\{i=a\}=p_a$, this equals
\begin{align}
\delta_{ab}\,p_a\ -\ p_a p_b\ -\ p_a p_b\ +\ p_a p_b
\ =\ \delta_{ab}\,p_a\ -\ p_a p_b,
\end{align}
i.e.
\begin{align}
\nabla^2 \mathcal{K}(0)\ =\ \mathrm{diag}(p)\ -\ p p^\top.
\end{align}

(iii) By Taylor’s theorem,
\begin{align}
\mathcal{K}(\Delta)\ =\ \tfrac{1}{2}\,\Delta^\top \big(\mathrm{diag}(p)-pp^\top\big)\,\Delta\ +\ o(\|\Delta\|^2).
\end{align}
Finally, note the algebraic identity (weighted variance):
\begin{align}
\Delta^\top\big(\mathrm{diag}(p)-pp^\top\big)\Delta
= \sum_{i=1}^n p_i \Delta_i^2 - \Big(\sum_{i=1}^n p_i \Delta_i\Big)^2
= \sum_{i=1}^n p_i\,(\Delta_i-\mu)^2,
\quad \mu:=\sum_{i=1}^n p_i \Delta_i.
\end{align}
\end{proof}

Now we prove \PRPref{main-kl-local}.  
Since the blockwise softmax-KL regularizer acts independently on each group $g$,
\begin{align}
\mathcal{K}_{\mathrm{blk}}(\Delta)
=\sum_{g=1}^G D_{\mathrm{KL}}\!\big(\softmax(v_{t-1}^{(g)}+\Delta^{(g)})\,\|\,\softmax(v_{t-1}^{(g)})\big),
\end{align}
where $\Delta^{(g)}=v_t^{(g)}-v_{t-1}^{(g)}$.
Applying \PRPref{kl-specific} to each group yields block Hessians
\begin{align}
H^{(g)}=\mathrm{diag}(p^{(g)})-p^{(g)}(p^{(g)})^\top,
\end{align}
which assemble into the block-diagonal
\begin{align}
H=\operatorname{blkdiag}(H^{(1)},\dots,H^{(G)}).
\end{align}
The variance identity holds within each group.

\section{Experiments}\label{app:experiments}

\subsection{Experimental Setup}\label{app:experimental_setup}

\begin{table}[]
  \centering\caption{Dataset statistics.}
  \resizebox{\linewidth}{!}{
\begin{tabular}{cl|rrrrrrr}
\hline
\multicolumn{1}{l}{} &  & Total Users & New Users & Total Items & New Items & Total Interactions & Avg Seq Len & Sparsity \\ \toprule
\multirow{6}{*}{Instruments} & $\CAL D_1$ & 17,046 & 17,046 & 40,471 & 40,471 & 141,788 & 8.32 & 0.9998 \\
 & $\CAL D_2$ & 1,772 & 1,183 & 8,346 & 2,900 & 13,197 & 7.45 & 0.9991 \\
 & $\CAL D_3$ & 1,821 & 1,265 & 8,325 & 2,909 & 13,334 & 7.32 & 0.9991 \\
 & $\CAL D_4$ & 2,289 & 1,684 & 9,617 & 3,864 & 18,811 & 8.22 & 0.9991 \\
 & $\CAL D_5$ & 2,238 & 1,699 & 9,131 & 3,365 & 17,573 & 7.85 & 0.9991 \\
\rowcolor{gray!20} & $\CAL D_{1:5}$ & 22,877 & NA & 53,509 & NA & 204,703 & NA & NA \\ \midrule
\multirow{6}{*}{Movies \& TVs} & $\CAL D_1$ & 17,928 & 17,928 & 39,228 & 39,228 & 190,411 & 10.62 & 0.9997 \\
 & $\CAL D_2$ & 1,866 & 1,141 & 11,612 & 1,479 & 17,665 & 9.47 & 0.9992 \\
 & $\CAL D_3$ & 2,106 & 1,200 & 12,658 & 1,926 & 19,874 & 9.44 & 0.9993 \\
 & $\CAL D_4$ & 2,284 & 1,357 & 13,788 & 1,882 & 22,929 & 10.04 & 0.9993 \\
 & $\CAL D_5$ & 2,332 & 1,552 & 13,491 & 1,559 & 22,225 & 9.53 & 0.9993 \\
 \rowcolor{gray!20}& $\CAL D_{1:5}$ & 23,178 & NA & 46,074 & NA & 273,104 & NA & NA \\ \midrule
\multirow{6}{*}{Books} & $\CAL D_1$ & 15,406 & 15,406 & 35,984 & 35,984 & 164,858 & 10.7 & 0.9997 \\
 & $\CAL D_2$ & 1,807 & 618 & 7,155 & 2,711 & 13,918 & 7.7 & 0.9989 \\
 & $\CAL D_3$ & 1,672 & 619 & 6,484 & 2,278 & 12,395 & 7.41 & 0.9989 \\
 & $\CAL D_4$ & 1,948 & 650 & 7,154 & 2,657 & 14,824 & 7.61 & 0.9989 \\
 & $\CAL D_5$ & 1,652 & 1,025 & 5,913 & 2,274 & 11,990 & 7.26 & 0.9988 \\
 \rowcolor{gray!20}& $\CAL D_{1:5}$ & 18,318 & NA & 45,904 & NA & 217,985 & NA & NA \\ \bottomrule
\end{tabular}}
\label{tab:datasets}
\end{table}

 \textbf{Datasets.}
We use the real-world temporal Amazon Review dataset, which contains user reviews (treated as implicit interactions) on Amazon products over time.\footnote{https://amazon-reviews-2023.github.io/} We focus on three categories: Musical Instruments, Movies \& TV, and Books. For Instruments and Movies \& TV, we use data from 2019–2023; for Books, we use 2022–2023. We take 60\% of the data as pretraining $\mathcal{D}_1$ and split the remaining 40\% into four equal incremental stages, $\CAL D_2,...,\CAL D_5$. For each incremental stage, we filter out users with fewer than five interactions. This ensures leave-one-out evaluation is feasible and makes incremental data even smaller than pretraining data, simulating real-world scenarios. Table~\ref{tab:datasets} summarizes dataset statistics, including the number of users, items, and interactions at each stage, average sequence length, and sparsity.

\textbf{Evaluation.}
For each $\mathcal{D}_t$, we apply leave-one-out per user: the second-to-last item is used for validation and the last item is reserved for testing. Following prior work~\citep{wang2024learnable, bao2025bi}, we construct multiple training pairs $(x_u,y_u)$ per user $u$ using a sliding window of size 20. The LLM trained on $\mathcal{D}_1$ serves as the pretrained model for all compared methods. At each stage $t=2,\dots,5$, after fine-tuning, the LLM autoregressively generates 10 items given the user history in the test pair. Generation uses constrained beam search restricted to valid item tokens, making it efficient and widely adopted in prior work~\citep{wang2024learnable, rajput2023recommender}. With these 10 items, we evaluate against the ground-truth item and report Hit@5, Hit@10, NDCG@5, and NDCG@10, averaged over $\mathcal{D}_2,\dots,\mathcal{D}_4$.

\textbf{Metrics.}
\textit{Hit@$k$} measures whether the ground-truth item appears among the $k$ generated items. For a user $u$ with ground-truth item $y_u$ and a ranked list of predictions $R_u$,
$$
\text{Hit@}k(u) = 
\begin{cases}
1 & \text{if } y_u \in R_u[1:k], \\
0 & \text{otherwise}.
\end{cases}
$$
\textit{NDCG@$k$} (Normalized Discounted Cumulative Gain) additionally accounts for the position of the ground-truth item, giving higher credit when it appears closer to the top:
$$
\text{NDCG@}k(u) = 
\begin{cases}
\frac{1}{\log_2(\text{rank}(y_u)+1)} & \text{if } y_u \in R_u[1:k], \\
0 & \text{otherwise},
\end{cases}
$$
Hit@$k$ captures whether the correct item is recommended at all, while NDCG@$k$ rewards ranking it higher in the list. We report averages of Hit@$k$ and NDCG@$k$ across all users, with $k \in {5,10}$.

\subsection{Learning Rate for Continual Stages}\label{app:learning_rate}
\begin{figure}
    \centering
    \includegraphics[width=0.4\textwidth]{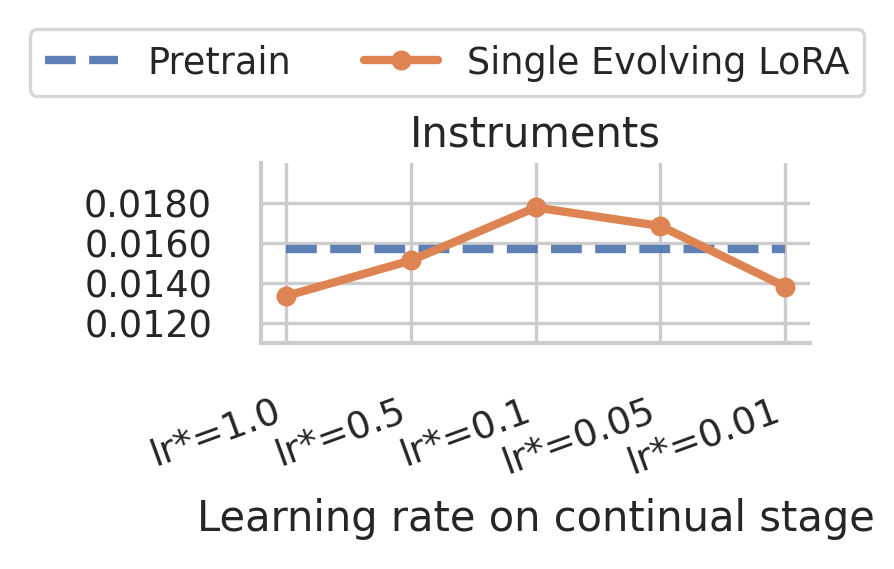} 
    \caption{Impact of the learning rate for continual data on model performance.}
    \label{fig:lr2}
\end{figure}

Incremental blocks are much smaller than the pretraining set $\mathcal{D}_1$ (see Appendix~\ref{app:experimental_setup}), making performance sensitive to the learning rate. Figure~\ref{fig:lr2} reports results for Single Evolving LoRA under different learning rates on incremental blocks. Using the pretraining rate (0.0002; $\texttt{lr*=}1.0$) performs worse than not updating on new data, likely due to overfitting. The best performance is achieved with $\texttt{lr*=}0.05$ or $\texttt{lr*=}0.1$, which aligns with the relative block size $|\mathcal{D}_t|/|\mathcal{D}_1|\approx 0.1$. This suggests that learning rates for incremental blocks should be scaled with respect to data size.



\begin{table}[]
\caption{Comparison of LLM-based and traditional methods in continual recommendation.}
\centering
\small
\begin{tabular}{cc|ccc}
\toprule
 &  & Instruments & Movies \& TVs & Books \\ \midrule
\multirow{5}{*}{Traditional two-tower} & Pretrain & 0.0153 & 0.0028 & 0.0041 \\
 & Fine-tuning & 0.0180 & 0.0114 & 0.0218 \\
 & Contrastive & 0.0177 & 0.0101 & 0.0272 \\
 & Contrastive + PIW & 0.0193 & 0.0113 & 0.0243 \\
 & PISA & 0.0194 & 0.0106 & 0.0301 \\ \midrule
\multirow{3}{*}{LLM-based } & Pretrain & 0.0157 & 0.0160 & 0.0235 \\
 & Fine-tuning (w/ LoRA) & 0.0178 & 0.0169 & 0.0414 \\
 & PESO & 0.0190 & 0.0173 & 0.0422 \\
 \bottomrule
\end{tabular}\label{tab:traditional}
\end{table}


\subsection{Comparison with Traditional Continual Recommender Systems}\label{app:traditional}
We compare our LLM-based methods (pretrain, single evolving LoRA, and \ours) against two-tower methods with LightGCN~\citep{he2020lightgcn} as backbone, including Pretrain, Fine-tuning, Contrastive~\citep{wang2021graph}, Contrastive+PIW~\citep{wang2023structure}, and PISA~\citep{yoo2025embracing}. Two-tower models use explicit user and item embeddings, and their continual methods mitigate forgetting by regularizing user embeddings against past versions: Contrastive maximizes mutual information between past and current embeddings, Contrastive+PIW further adapts the regularization weights per user, and PISA combines stability and plasticity regularization.

Table~\ref{tab:traditional} reports results averaged across time stages and metrics. First, LLM-based recommenders (both pretrain and continual) generally outperform traditional methods, highlighting the generalization ability and knowledge transfer benefits of LLMs. On Instruments, however, the performance gap is smaller, suggesting that explicit dual modeling of users and items still provides benefits for capturing collaborative signals. It is worth noting that there also remains considerable headroom for LLM-based models if larger beam sizes are used during generation.

Second, While \ours\ outperforms traditional continual methods in absolute terms, the relative gains of continual techniques over their respective pretraining baselines are larger in traditional settings. This is likely because two-tower methods explicitly capture preference shifts through user embeddings, supporting our view that modeling user preference drift is crucial in continual recommendation. At the same time, it underscores the difficulty of capturing such dynamics in LLM-based methods, pointing to an important direction for future research.

\subsection{Quantification of Distribution Shift}
\label{app:distribution_shift}
To validate the realism of our continual learning formulation, we explicitly quantified the user preference drift between data blocks on the Instruments dataset. We employed a domain discrimination approach:
\begin{enumerate}
    \item We embed user interaction sequences into fixed-dimensional vectors using pretrained codebooks.
    \item For each pair of blocks $(t-1)$ and $(t)$, we train a binary logistic regression classifier to distinguish samples from the two blocks.
    \item We compute the \textbf{Drift Score} $\delta(t-1,t) = 2(\text{AUC}-0.5) \in [0,1]$, where 0 implies identical distributions and 1 implies completely separable distributions.
\end{enumerate}

Table~\ref{tab:drift_score} reports both the step-wise drift $\delta(t-1,t)$ and the cumulative drift from the base block $\delta(0,t)$. The results show non-trivial step-wise drift and, crucially, a steady increase in cumulative drift (reaching 0.457 at $t=4$). This confirms that user preferences structurally evolve away from the initial state, validating our experimental setup.

\begin{table}[h]
    \centering
    \caption{Quantification of distribution shift (Drift Score) on the Instruments dataset.}
    \label{tab:drift_score}
    \begin{tabular}{lcccc}
        \toprule
        Measure & $t=1$ & $t=2$ & $t=3$ & $t=4$ \\
        \midrule
        Step-wise $\delta(t-1,t)$ & 0.200 & 0.060 & 0.240 & 0.090 \\
        Cumulative $\delta(0,t)$  & 0.200 & 0.311 & 0.342 & 0.457 \\
        \bottomrule
    \end{tabular}
\end{table}

\subsection{Comparison with Additional Baselines}
\label{app:additional_baselines}

\subsubsection{Comparison with Standard Training Paradigms (Full-Parameter Fine-Tuning \& Retraining)}
To validate the effectiveness of our LoRA-based sequential fine-tuning approach, we compared it against two traditional training paradigms:
\begin{enumerate}
    \item \textbf{Full-Parameter Fine-Tuning:} Updating \textit{all} LLM parameters sequentially on new data blocks.
    \item \textbf{Retraining (with LoRA):} 
    At each stage $t$, we restart from the same pretrained LLM (i.e., without warm-starting from stage $t{-}1$) and fine-tune it on the \textit{cumulative dataset} $(\mathcal D_1 \cup \dots \cup \mathcal D_t)$. 
\end{enumerate}

As shown in Table~\ref{tab:standard_training}, \textbf{Single Evolving LoRA} consistently outperforms both approaches.
\begin{itemize}
    \item \textbf{vs. Full-Parameter Fine-Tuning:} Full-parameter updates suffer from a dilemma: high learning rates ($2\mathrm{e}{-5}$) lead to catastrophic forgetting, while low rates ($2\mathrm{e}{-6}$) result in insufficient adaptation. LoRA acts as a structural regularizer, mitigating forgetting while enabling effective adaptation.
    \item \textbf{vs. Retraining:} While full retraining outperforms static pretraining, it underperforms sequential fine-tuning. This aligns with prior work~\citep{yoo2025embracing}, suggesting that sequential updates naturally prioritize recent preference signals, whereas full retraining treats old and new data equally, diluting the signal of evolving interests.
\end{itemize}

\begin{table}[h]
    \centering
    \caption{Comparison with Standard Training Paradigms (Full Fine-Tuning and Full Retraining).}
    \label{tab:standard_training}
    \begin{tabular}{lcccc}
        \toprule
        Method & Hit@5 & Hit@10 & NDCG@5 & NDCG@10 \\
        \midrule
        Pretrain (Static) & 0.0166 & 0.0216 & 0.0115 & 0.0131 \\
        Full Retraining (Cumulative Data) & 0.0170 & 0.0231 & 0.0121 & 0.0141 \\
        \midrule
        Full Fine-Tuning ($lr = 2\mathrm{e}{-5}$) & 0.0142 & 0.0228 & 0.0099 & 0.0127 \\
        Full Fine-Tuning ($lr = 2\mathrm{e}{-6}$) & 0.0171 & 0.0254 & 0.0122 & 0.0149 \\
        \midrule
        \textbf{Single Evolving LoRA} (LoRA Fine-Tuning) & \textbf{0.0181} & \textbf{0.0253} & \textbf{0.0127} & \textbf{0.0150} \\
        \bottomrule
    \end{tabular}
\end{table}

\subsubsection{Comparison with Additional Continual LoRA Methods}

We compared \ours\ against O-LoRA~\citep{wang2023orthogonal}, AM-LoRA~\citep{liu2024learning}, and LSAT~\citep{shi2024preliminary} on the Instrument dataset. O-LoRA and AM-LoRA belong to the cumulative family but use \textit{orthogonality} or \textit{attention mechanisms} to combine adapters, while LSAT utilizes \textit{adapter interpolation}. As shown in Table~\ref{tab:peft_comparison}, \ours\ consistently outperforms all of them. This supports our claim that explicitly maintaining discrete adapters is less effective for gradual preference drift than our proximal regularization approach.

\begin{table}[h]
    \centering
    \caption{Comparison with recent Continual PEFT methods on Instruments.}
    \label{tab:peft_comparison}
    \begin{tabular}{lcccc}
        \toprule
        Method & Hit@5 & Hit@10 & NDCG@5 & NDCG@10 \\
        \midrule
        Single Evolving LoRA & 0.0181 & 0.0253 & 0.0127 & 0.0150 \\
        Cumulative LoRA & 0.0182 & 0.0260 & 0.0129 & 0.0154 \\
        \midrule
        O-LoRA & 0.0191 & 0.0259 & 0.0134 & 0.0156 \\
        AM-LoRA & 0.0182 & 0.0240 & 0.0125 & 0.0144 \\
        LSAT & 0.0164 & 0.0250 & 0.0117 & 0.0144 \\
        LSAT (+ Param Inheritance) & 0.0183 & 0.0254 & 0.0130 & 0.0153 \\
        \midrule
        \textbf{\ours} & \textbf{0.0193} & \textbf{0.0268} & \textbf{0.0138} & \textbf{0.0162} \\
        \bottomrule
    \end{tabular}
\end{table}

\subsection{Performance on LC-REC Backbone}
To demonstrate robustness across architectures, we evaluated \ours\ using the LC-REC backbone~\citep{zheng2024adapting}. As shown in Table~\ref{tab:lcrec}, \ours\ maintains its superiority over baselines.

\begin{table}[h]
    \centering
    \caption{Performance comparison using the LC-REC backbone.}
    \label{tab:lcrec}
    \begin{tabular}{lcccc}
        \toprule
        Method & Hit@5 & Hit@10 & NDCG@5 & NDCG@10 \\
        \midrule
        Single Evolving LoRA & 0.0164 & 0.0249 & 0.0119 & 0.0146 \\
        Cumulative LoRA & 0.0178 & 0.0249 & 0.0122 & 0.0145 \\
        SD-LoRA & 0.0185 & 0.0256 & 0.0127 & 0.0150 \\
        \textbf{\ours} & \textbf{0.0179} & \textbf{0.0266} & \textbf{0.0130} & \textbf{0.0158} \\
        \bottomrule
    \end{tabular}
\end{table}

\subsection{Performance on Non-E-Commerce Dataset (Yelp)}
To further explore non-e-commerce domains, we evaluated \ours\ on the Yelp dataset, where interactions correspond to user check-ins at locations, using the same data-splitting strategy as in our main experiments. As shown in Table~\ref{tab:yelp_results}, \ours\ consistently outperforms strong competitors, including Single Evolving LoRA and SD-LoRA.

This result is particularly notable given that, unlike Amazon products which feature detailed textual descriptions, Yelp locations often lack deep semantic content (consisting primarily of names like ``Pizza Hut'' or coarse categories like ``Pizza'' or ``Restaurant''). This demonstrates the robustness of our method, showing it remains highly effective even in settings with limited semantic richness.

\begin{table}[h]
    \centering
    \caption{Performance comparison on the Yelp dataset.}
    \label{tab:yelp_results}
    \begin{tabular}{lcccc}
        \toprule
        Methods & Hit@5 & Hit@10 & NDCG@5 & NDCG@10 \\
        \midrule
        Pretrain & 0.0201 & 0.0309 & 0.0126 & 0.0161 \\
        Single Evolving LoRA & 0.0290 & 0.0442 & 0.0190 & 0.0239 \\
        SD-LoRA & 0.0279 & 0.0432 & 0.0168 & 0.0230 \\
        \textbf{\ours} & \textbf{0.0302} & \textbf{0.0454} & \textbf{0.0199} & \textbf{0.0248} \\
        \bottomrule
    \end{tabular}
\end{table}

\subsection{Explicit Measurement of Forgetting}
\label{app:forgetting}
We measured the performance drop on past blocks to analyze forgetting behavior. Table~\ref{tab:forgetting} shows the difference between the final model's performance on $D_t$ and its initial performance at time $t$. While \ours\ shows selective forgetting on intermediate blocks (allowing it to shed obsolete trends), it achieves the highest overall performance and best retrieval for dormant users, indicating that this forgetting is benign and adaptive rather than catastrophic.

\begin{table}[h]
    \centering
    \caption{Performance drop on past blocks (lower is strictly less forgetting, but may imply rigidity).}
    \label{tab:forgetting}
    \begin{tabular}{lcccc}
        \toprule
        Method & Drop on $D_0$ & Drop on $D_1$ & Drop on $D_2$ & Drop on $D_3$ \\
        \midrule
        Single Evolving LoRA & 0.0062 & 0.0087 & 0.0042 & 0.0031 \\
        Cumulative LoRA & 0.0060 & 0.0062 & 0.0035 & 0.0060 \\
        \textbf{\ours} & 0.0062 & 0.0107 & 0.0048 & 0.0045 \\
        \bottomrule
    \end{tabular}
\end{table}


\section{Efficiency Analysis}
\label{app:efficiency}
\ours\ introduces negligible overhead compared to baselines:
\begin{itemize}
    \item \textbf{Storage Complexity:} \ours\ stores only one previous LoRA adapter, resulting in $O(1)$ storage complexity relative to the number of stages. In contrast, Cumulative LoRA grows linearly $O(T)$ as it must store all past adapters.
    \item \textbf{Computational Complexity:} \ours\ adds only a lightweight quadratic/KL penalty to the loss. This requires no additional forward passes. In practice, we observed no measurable slowdown in training time compared to standard Single LoRA fine-tuning.
\end{itemize}

\section{Discussion on Prompt Tuning vs. LoRA}
Prompt-tuning–based PEFT methods typically learn a prompt pool and dynamically retrieve the most relevant prompts for each input, inserting them into the input or intermediate representations without updating backbone weights~\citep{wang2022learning} This introduces inference overhead because the model must compute query features and perform similarity matching over a growing prompt pool at inference time.
The inference-inefficiency is even more severe in our generative recommendation setting: autoregressive generation requires many forward passes per prediction, and each step would need repeated prompt retrieval. Recent studies in vision~\citep{wu2025sd, liang2024inflora} also report that LoRA-based methods generally outperform prompt-based approaches in large-scale tasks, making LoRA the preferred PEFT technique.

\section{Additional Related Work}\label{app:additional_related_work}
In modern machine learning research---including recommender systems, personalized LLM agents, and memory~\citep{wei2025evo,wei2026agentic,bei2026mem,ningmc,liu2025selfelicit,liu2025seeing,liu2025breaking,liu2024class,li2025graph,li2025language,lin2024backtime,qiu2025efficient,qiu2025ask,qiu2025efficient2,he2026fedecider,lee2025sprint,xu2024discrete}---temporal adaptation under distribution shift~\citep{zeng2026pave} has been extensively studied. Our work focuses on parameter-efficient tuning and regularization/distillation to mitigate forgetting and lock-in in continual recommendation settings. It complements this line of research by maintaining a single evolving LoRA adapter and applying a proximal regularizer that provides data-aware control of the stability--plasticity trade-off.

\section{Prompt}
We show below the template used in all experiments. Notably, \texttt{<a\_[i1]><b\_[j1]><c\_[k1]><d\_[l1]>} represents one
user–item interaction encoded as four semantic-ID tokens. For instance, \texttt{<a\_144><b\_72><c\_103><d\_217>} is one such
tuple describing a single interacted item~\citep{rajput2023recommender, wang2024learnable}.
\begin{promptframe}
\begin{lstlisting}[basicstyle=\ttfamily\small]
Below is an instruction that describes a task. 
Write a response that appropriately completes the request.\n\n

### Instruction:\n
Based on the items that the user has interacted with: 
<a_[i1]><b_[j1]><c_[k1]><d_[l1]>,
<a_[i2]><b_[j2]><c_[k2]><d_[l2]>,
...,
<a_[iN]><b_[jN]><c_[kN]><d_[lN]>,
can you determine what item would be recommended to the user next?\n\n
### Response:
\end{lstlisting}
\end{promptframe}

\section{Use of Large Language Models}
LLMs were used only for writing polish (grammar and clarity). All content was reviewed and approved by the authors. LLMs did not contribute to research ideation, algorithm design, implementation, or analysis.

\end{document}